\documentclass[10pt, conference, compsocconf]{IEEEtran}
\pdfoutput=1

\ifCLASSINFOpdf
   \usepackage[pdftex]{graphicx}
\else
   \usepackage[dvips]{graphicx}
\fi
\usepackage{amsmath}
\usepackage{amssymb}

\usepackage{algorithmic}
\usepackage{algorithm}
\usepackage{adjustbox}

\usepackage{balance}
\usepackage{flushend}

\usepackage{array}

\usepackage[caption=true]{caption}
\usepackage[font=footnotesize]{subfig}
\usepackage{fixltx2e}

\usepackage{stfloats}

\usepackage{url}

\renewcommand{\algorithmicrequire}{\textbf{Input:}}
\renewcommand{\algorithmicensure}{\textbf{Output:}}

\begin{document}
%
\title{Error-Robust Multi-View Clustering}


\author{\IEEEauthorblockN{Mehrnaz Najafi}
\IEEEauthorblockA{Department of Computer Science\\
University of Illinois at Chicago\\
Chicago, IL, USA\\
mnajaf2@uic.edu}
\and
\IEEEauthorblockN{Lifang He}
\IEEEauthorblockA{Weill Cornell Medicine College\\
Cornell University\\
New York, NY, USA\\
lifanghescut@gmail.com}
\and
\IEEEauthorblockN{Philip S. Yu}
\IEEEauthorblockA{Department of Computer Science\\
University of Illinois at Chicago\\
Chicago, IL, USA\\
psyu@uic.edu}
}


%


\maketitle

\begin{abstract}
In the era of big data, data may come from multiple sources, known as multi-view data. 
Multi-view clustering aims at generating better clusters by exploiting complementary and consistent information from multiple views rather than relying on the individual view. 
Due to inevitable system errors caused by data-captured sensors or others, the data in each view may be erroneous. 
Various types of errors behave differently and inconsistently in each view. 
More precisely, error could exhibit as noise and corruptions in reality. 
Unfortunately, none of the existing multi-view clustering approaches handle all of these error types. 
Consequently, their clustering performance is dramatically degraded. 
In this paper, we propose a novel Markov chain method for Error-Robust Multi-View Clustering (EMVC). 
By decomposing each view into a shared transition probability matrix and error matrix and imposing structured sparsity-inducing norms on error matrices, we characterize and handle typical types of errors explicitly. 
To solve the challenging optimization problem, we propose a new efficient algorithm based on Augmented Lagrangian Multipliers and 
prove its convergence rigorously. 
Experimental results on various synthetic and real-world datasets show the superiority of the proposed EMVC method over the baseline methods 
and its robustness against different types of errors.

\end{abstract}


%
\IEEEpeerreviewmaketitle

\section{Introduction}
In the era of big data, data may have different views (i.e., variety), where observations are represented by multiple sources, known as \textit{multi-view data}. For instance, 
specific news may be available on different broadcasting websites such as BBC and CNN so that each website represents a view of the same news.
As another example, in Wikipedia, concept of dog may have multiple representations in the form of image and text. 
Fig. 1 shows the mentioned examples of multi-view data. 

\begin{figure}[h]
\centering
\includegraphics[width=5.5cm]
{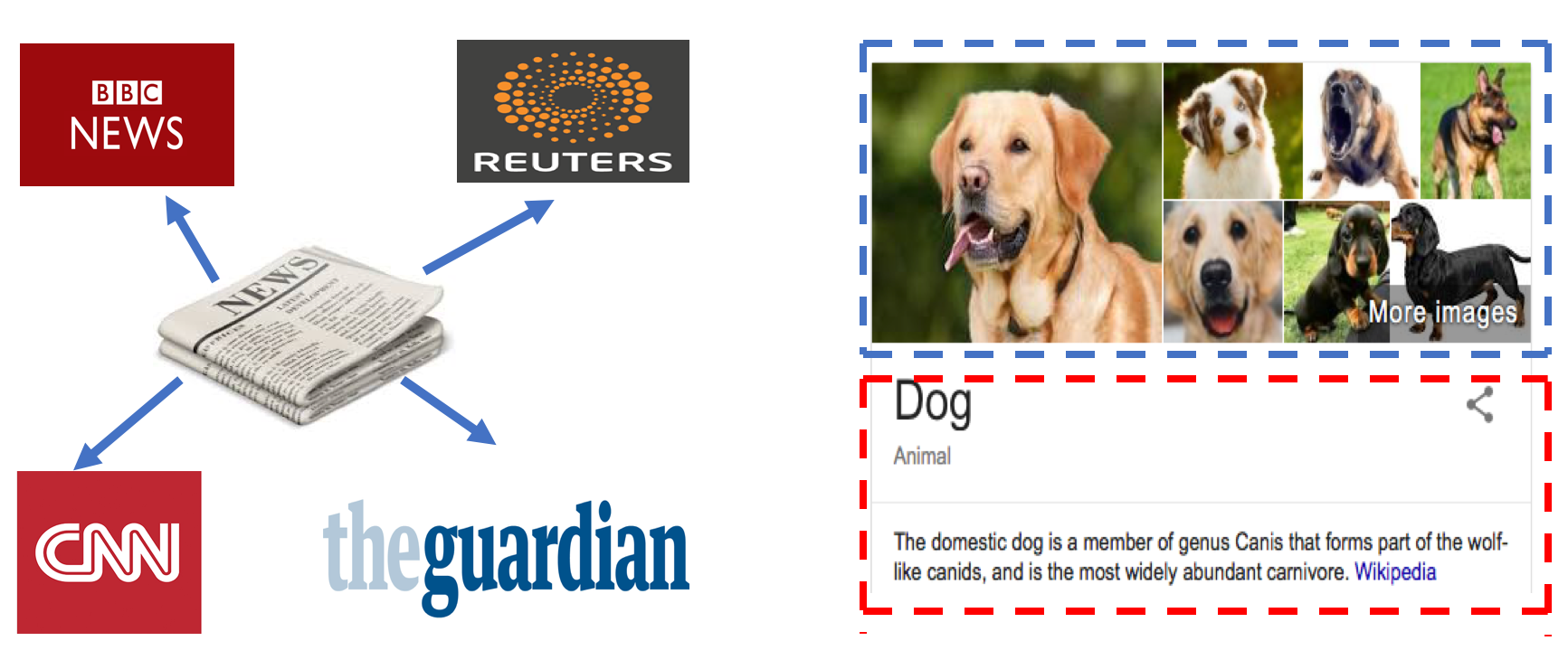}
\caption{Multi-View data}
\label{fig:multi-data}
\end{figure}

Multi-view data commonly have the following properties:
\begin{itemize}
    \item Each view may be represented by an arbitrary type and number of features. It may be also collected from diverse domains (variety of big data). Different views often contain \textit{complementary} and \textit{compatible} information to each other \cite{zhao2017multi}. 
    For instance, in Wikipedia, one view (image) consists of vision features, while another view (text) has textual features.
    
    \item Due to inevitable system errors caused by data extractors, each view may be \textit{erroneous} (veracity of big data). Generally, error refers to the deviation between model assumption and data. It could exhibit as \textit{noise} and \textit{corruptions} in reality \cite{Liu13}.  
    Fig. 2 illustrates three types of errors. Noise refers to \textit{slight} perturbation of random subset of entries in data. Random corruptions indicate that a fraction of random entries are \textit{grossly} perturbed, while sample-specific corruptions (or outliers) represent the phenomena that a fraction of the samples (or data points) in each view are far away from the real values. Real-world multi-view data can encounter any or combination of these error types.
    
    
\end{itemize}

\begin{figure}[h]
        \centering
        \subfloat[]{
                \centering
                \includegraphics[width=0.24\linewidth]{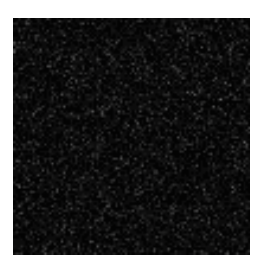}
                \label{fig:noise}\hfill}
        \subfloat[]{
                \centering
                \includegraphics[width=0.22\linewidth]{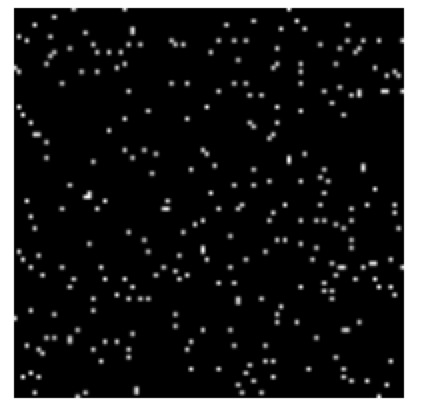}
                \label{fig:random-cor}\hfill}
        \subfloat[]{
                \centering
                \includegraphics[width=0.23\linewidth]{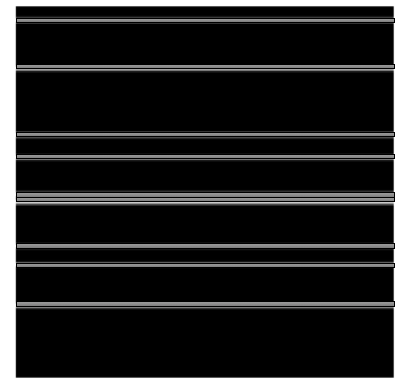}
                \label{fig:sample-cor}\hfill}
    \label{fig:corruptions}
    \caption{Three types of errors (what we show is a perturbed data matrix whose rows are samples and columns are features). (a) Noise (b) Random Corruptions (c) Sample-Specific Corruptions (or Outliers)}
 \end{figure}

\vspace{-0.3cm}

Due to ever increasing need of learning from multi-view data and lack of label information in many applications, clustering on multiple views has received considerable attention recently \cite{Bickel:2004:MC:1032649.1033432,desa05spectral,zhao2017multi}. The problem is referred to as \textit{multi-view clustering} aiming at finding compatible clusters of data points across all views. 
One of the most common algorithms used for multi-view clustering is \textit{spectral clustering} \cite{Ng:2001:SCA:2980539.2980649}. 
 
 
Spectral clustering models the data which is in the form of a graph where nodes are data points and edges represent similarities between data points \cite{Ng:2001:SCA:2980539.2980649}. First, it projects all the data points to a new low-dimensional space where they are easily separable. The new space is built with the eigen-decomposition of the Laplacian matrix of the graph. Then, it finds the clusters by applying another clustering algorithm like $k$-means. It is shown that spectral clustering can find the normalized min-cut of the graph \cite{Ng:2001:SCA:2980539.2980649}. Also, there is a hidden relation between spectral clustering and \textit{Markov chains} \cite{Shi:2000:NCI:351581.351611}. In single-view clustering, Laplacian of the graph can be obtained by real relaxation of the combinatorial normalized cut \cite{Xia:2014:RMS:2892753.2892850}. Then, it is converted to a transition probability matrix which generates a Markov chain on the graph \cite{Xia:2014:RMS:2892753.2892850}. In the context of multi-view clustering, the transition probability matrix would need to get built across all views. 


A challenging problem may arise when the views are erroneous, which makes the corresponding transition probability matrices being perturbed. It could then result in portion of data points being assigned to wrong clusters. 
To address multi-view clustering with noisy views, Xia \emph{et al} proposed a method, named as RMSC 
\cite{Xia:2014:RMS:2892753.2892850}. This approach decomposes a transition probability matrix of each view into two parts: a shared transition probability matrix across all views and an error matrix which encodes the noise in the transition probability matrix in each view. The error matrix of each view captures the difference between the transition probabilities of that view and their correspondings in the shared transition probability matrix. RMSC assumes sparse representation error matrix via $\ell_1$ norm.

One of the shortcomings of RMSC is that it \textit{only} handles noise in data. Specifically, since it only imposes $\ell_1$ norm on the error matrices, it cannot deal with sample-specific corruptions well. This is because error matrix with sample-specific corruptions has sparse row supports. Also, RMSC treats each error matrix independently. However, data may come from various sources, which could result in error matrices with inconsistent magnitude values, and thus degradation of clustering performance, when error matrices are treated independently \cite{Liu13}. 



To handle typical types of errors in multi-view data, we propose a novel Error-Robust Multi-View Clustering (EMVC) method based on Markov chains. Different from RMSC \cite{Xia:2014:RMS:2892753.2892850}, EMVC is based on integration of low-rank decomposition and \textit{group} $\ell_1$ \cite{huang09} and $\ell_{2,1}$ regularization terms, aiming to learn a shared transition probability matrix where the transition probability matrices of different views will be co-regularized to a common consensus, while improving the robustness of clustering. 

In some cases, features of a certain view are more or less discriminative for clusters. As a result, error in more discriminative features could substantially decrease clustering performance. To improve robustness of clustering on this error, we impose \textit{group} $\ell_1$ norm on error matrix. In this way, in contrast to $\ell_1$ norm, group $\ell_1$ norm learns group-wise features importance of one view on each cluster and thus improves robustness of clustering against erroneous discriminative features. 
By various experiments, we also claim that by using \textit{group} $\ell_1$ rather than $\ell_1$ norm, the proposed EMVC method achieves better clustering performance both on non-erroneous and erroneous datasets.

To deal with sample-specific corruptions, EMVC imposes $\ell_{2,1}$ norm on error matrix because similar to \cite{Liu13}, error matrix with this error type has sparse row supports. 
Furthermore, since data may come from multiple heterogeneous sources, error matrices could have inconsistent magnitude values. In contrast to RMSC, as suggested in \cite{Liu13}, with the aim of increasing clustering performance, we enforce the column of each error matrix with respect to each view to have jointly consistent magnitude values by vertical concatenation of error matrices of all views.

The $\ell_{2,1}$ and group $\ell_1$ norms are two non-smooth structured sparsity-inducing norms, which make the corresponding objective function of EMVC challenging to optimize. We present a reformulation of the objective function and propose a new efficient optimization algorithm based on the Augmented Lagrangian Multiplier \cite{LinChenMa2010} to optimize it. We also present a rigorous proof of convergence for the optimization technique.
Our contributions can be summarized as follows:
\begin{enumerate}
    \item To the best of our knowledge, EMVC is the first work that 
    can address any or combination of the typical types of errors in multi-view clustering via combination of $\ell_{2,1}$ and group $\ell_1$ norms. 
    Since it is generally hard to know which type of error incurred in each view, it is important to have an all-encompassing approach that can handle any or combination of the typical error types.
    
    
    \item EMVC is the first Markov chains method that does not treat error matrices independently. Independent treatment of error matrices could decrease clustering performance \cite{Liu13}. The proposed EMVC method enforces the error matrices in each view to have jointly consistent magnitude values. 
    
    \item We propose a new efficient optimization algorithm to solve the EMVC optimization problem, along with rigorous proof of convergence.
    
    \item Through extensive experiments on synthetic and real-world datasets, we show that EMVC is superior to several state-of-the-art methods in the multi-view clustering and robust against typical error types.
    

\end{enumerate}
\section{Preliminaries}
In this section, we introduce some related concepts and notations. The mathematical notations used in the rest of the paper are summarized in Table \ref{tab:notation}.

\subsection{Transition Probability Matrix}
Given a graph $G$ with $N$ nodes, a square matrix called transition probability matrix is defined over $G$ that contains the transitions of a Markov chain. It is denoted as $\mathbf{P} \in \mathbb{R}^{N \times N}$. Each element of $\mathbf{P}$ denotes a probability (i.e., $p_{i,j} >= 0$), and all outgoing transitions from a specific state have to sum to one (i.e., $\sum_{j = 1}^{} p_{i, j} = 1$). Each row in $\mathbf{P}$ is a distribution probability over the transitions of the corresponding state. 

\begin{table}
\centering
\caption{List of basic symbols}
\label{tab:notation}
\begin{tabular}{c l}
\hline
\textbf{Symbol} & \textbf{Definition and description}\\
\hline
$x$ & each lowercase letter represents a scale\\ 
$\mathbf{x}$ & each boldface lowercase letter represents a vector\\
$\mathbf{X}$ & each boldface uppercase letter represents a matrix\\ 
$\left\langle\cdot,\cdot\right\rangle$ & denotes inner product\\
$rank(\cdot)$ & denotes the rank of the matrix\\
$Tr(\cdot)$ & denotes the trace of the matrix\\
\hline
\end{tabular}
\end{table}


\subsection{Spectral Clustering via Markov chains}
Spectral clustering seeks clusters of data points in a weighted graph $G$ where vertices are data points and edges represent similarity between two connecting data points. There is a relationship between spectral clustering and transition probability matrix \cite{Shi:2000:NCI:351581.351611}. Spectral clustering on graph $G$ is equivalent to finding clusters on $G$ such that the Markov random walk remains long within the same cluster and jumps infrequently between clusters \cite{Xia:2014:RMS:2892753.2892850}.

In the context of clustering, a natural way to construct a transition probability matrix is to first build a similarity matrix $\mathbf{S}$ between pairs of data points and then calculate the corresponding transition probability matrix $\mathbf{P}$ by $\mathbf{P} = (\mathbf{D}^{-1} \mathbf{S})$, where $\mathbf{D}$ denotes the degree matrix of graph $G$. One way to build a similarity matrix $\mathbf{S}$ is to use Gaussian kernels \cite{Xia:2014:RMS:2892753.2892850}. Let $s_{i,j}$ denotes the similarity on a pair of data points $\mathbf{x}_{i}$ and $\mathbf{x}_{j}$. It can be calculated as follows:
\begin{equation}
{\small
s_{i,j} = exp(- ||\mathbf{x}_{i} - \mathbf{x}_{j}||^{2}_{2} /\sigma^{2})
}
\label{for:1}
\end{equation}
\noindent where $||.||_{2}$ denotes the $l_{2}$-norm and $\sigma^{2}$ indicates the standard deviation (e.g., it could be set to median of Euclidean distance over all pairs of data points). Algorithm \ref{TPM} summarizes the overall scheme for computing transition probability matrix.

\begin{algorithm}
\caption{Transition Probability Matrix Construction}
\label{TPM}
\begin{algorithmic} [1]
\REQUIRE Data matrix $\mathbf{X} \in \mathbb{R}^{N \times D}$
\ENSURE Transition probability matrix $\mathbf{P} \in \mathbb{R}^{N \times N}$\\
\FOR{$i = 1, ..., N$}
\FOR{$j = 1, ..., N$}
\STATE $s_{i, j} = exp(- ||\mathbf{x}_{i} - \mathbf{x}_{j}||^{2}_{2} /\sigma^{2})$
\ENDFOR
\ENDFOR
\FOR{$i = 1, ..., N$}
\STATE $d_{i, i} = \sum_{j = 1}^{N} s_{i, j}$
\ENDFOR
\STATE $\mathbf{P} = \mathbf{D}^{-1} \mathbf{S}$
\end{algorithmic}
\end{algorithm}

The steps of spectral clustering via Markov chains is described in Algorithm \ref{spec-markov} \cite{Zhou:2005:LLU:1102351.1102482}. To perform clustering using Markov chains, a crucial step is to build the transition probability matrix $\mathbf{P}$ over graph $G$ in Line 1. A stationary distribution of $\mathbf{P}$ is obtained in Line 2. Two matrices $\mathbf{L}$ and $\hat{\mathbf{D}}$ are computed in Lines 3 and 4. Finally, \textit{k}-means must be performed on eigenvectors of the generalized eigenproblem $\mathbf{L} \mathbf{u} = \lambda \hat{\mathbf{D}}\mathbf{u}$ (Lines 5 and 6).


 
 
 
 
 

\begin{algorithm}
\caption{Spectral Clustering via Markov Chains}
\label{spec-markov}
\begin{algorithmic}[1]
\REQUIRE Graph $G$
\ENSURE Clustering results \\

 \STATE Define a random walk over $G$ with a transition probability matrix $\mathbf{P}$ constructed by Algorithm \ref{TPM}

 \STATE Compute stationary distribution $\pi$ satisfying $\pi = \mathbf{P} \pi$
 
 \STATE Build diagonal matrix $\hat{\mathbf{D}}$ with the $i$-th diagonal element as $\hat{d}_{i,i}$, such that $\hat{d}_{i,i} = \pi(i)$
 
 \STATE $\mathbf{L} = \hat{\mathbf{D}}  - ((\hat{\mathbf{D}} \mathbf{P} + \mathbf{P}^{T} \hat{\mathbf{D}} )/2)$
 
 \STATE Calculate $R$ smallest generalized eigenvectors $\mathbf{u}_{1}, ..., \mathbf{u}_{r}$ of the generalized eigenproblem $\mathbf{L} \mathbf{u} = \lambda \hat{\mathbf{D}} \mathbf{u}$
 
 \STATE Cluster $\mathbf{U}$ by $k$-means to obtain clustering results, where $\mathbf{U}$ is the matrix consisting of the vectors $\mathbf{u}_{1}, ..., \mathbf{u}_{r}$.
\end{algorithmic}
\end{algorithm}
\renewcommand{\algorithmicrequire}{\textbf{Input:}}
\renewcommand{\algorithmicensure}{\textbf{Output:}}
\section{Error-Robust Multi-View Clustering}

In this section, we will first systematically propose a novel error-robust multi-view algorithm for clustering, followed by a new efficient iterative algorithm to solve the formulated non-smooth objective function. 

\textbf{Problem.} In the setting of clustering, given $N$ distinct data points or samples with $K$ related views, their views are denoted as $\mathbf{X}^{(1)}, \mathbf{X}^{(2)}, ..., \mathbf{X}^{(K)}$. The goal is to derive a clustering solution across all views. We assume that the features in each individual view are sufficient for obtaining most of the clustering information and each individual view might be erroneous. 

\subsection{Transition Probability Matrix Construction}

Transition probability matrices have been used to model multiple views \cite{Xia:2014:RMS:2892753.2892850,Zhou:2007:SCT:1273496.1273642}.
There are several ways to build a transition probability matrix with respect to each view. 
We use Algorithm \ref{TPM} to construct transition probability matrices with respect to each individual view.

\subsection{Problem Formulation}
\label{subsec:pf}
Assuming that each individual view might be erroneous so that it causes wrong assignment of data points to clusters, each transition probability matrix $\mathbf{P}^{(k)}$ can be decomposed into two terms: a shared transition probability matrix $\mathbf{\hat{P}}$ and the error matrix $\mathbf{E}^{(k)}$ that indicates the error in the transition probabilities in view $k$.
\begin{equation}
\forall k, \mathbf{P}^{(k)} = \mathbf{\hat{P}} + \mathbf{E}^{(k)}
\label{for:2}
\end{equation}
We use $\mathbf{\hat{P}}$ as the input transition probability matrix to the Markov chains method (i.e., Algorithm 2) to obtain clustering solution. Using the transition probability construction method described in Algorithm \ref{TPM}, we get initial transition probability matrices with respect to each individual view. 

In order to approximate the shared transition probability matrix while reducing its complexity, we minimize $rank(\mathbf{\hat{P}})$ (i.e., low rankness criterion). 
Since data may come from different sources, error matrices could have inconsistent magnitude values, which adversarially affects clustering performance \cite{Liu13}. To enforce the column of $\mathbf{E}^{(k)}$ in each view to have jointly consistent magnitude values, we vertically concatenate error matrices of views along their columns (i.e., $\mathbf{E} = [\mathbf{E}^{(1)}; \mathbf{E}^{(2)}; ...; \mathbf{E}^{(K)}]$).

In some cases, error may appear in features of a specific view which are more or less  discriminative for clustering. Compared to error in less discriminative features, error in more discriminative features degrade clustering performance significantly. 
For example, color features substantially affects the detection of traffic light and trees whereas they are irrelevant for finding cars in the context of image clustering. Thus, error in color features would substantially decrease clustering performance. 
To improve robustness of clustering against error in these features, we add group $\ell_{1}$ norm \cite{huang09} on $\mathbf{E}$. The group $\ell_{1}$ norm can be defined as follows:
\begin{align}
& ||\mathbf{E}||_{G1} = \sum_{i = 1}^{N} \sum_{j = 1}^{K} ||\mathbf{e}_{i}^{j}||_{2}
\quad \text{s.t.} \hspace{0.3cm} \mathbf{E} = [\mathbf{E}^{(1)}; \mathbf{E}^{(2)}; ...; \mathbf{E}^{(K)}]
\label{for:l1}
\end{align}
\noindent where $\mathbf{e}_{i}^{j}$ denotes $\mathbf{E}((j-1) \times N+1:j \times N, i)$ (i.e., the segment $((j-1) \times N+1:j \times N)$ of $i$-th column of $\mathbf{E}$). 
This norm uses $\ell_{2}$ norm within each view and $\ell_1$ norm between views. Thus, it enforces the sparsity between different views, i.e., if features of one view are not discriminative for clustering, Eq. (3) will assigns zero to them.

In sample-specific corruptions, $\mathbf{E}$ has sparse row supports \cite{Liu13}. 
Thus, to handle error in specific samples, we add $\ell_{2,1}$ penalty \cite{nie2010efficient} on $\mathbf{E}$. The $\ell_{2,1}$ norm is defined as follows:
\begin{equation}
||\mathbf{E}||_{2,1} = \sum_{i = 1}^{K \times N} ||\mathbf{e}^{i}||_{2}
\label{for:l21}
\end{equation}

\noindent where $\mathbf{e}^{i}$ denotes $\mathbf{E}(i,:)$ (i.e., the $i$-th row of $\mathbf{E}$). 


Based on the above consideration, the objective function for obtaining a shared transition probability matrix is formulated as follows:
\begin{align}
\label{for:4}
& \min_{\mathbf{\hat{P}}, \mathbf{E}} rank(\mathbf{\hat{P}}) + \beta ||\mathbf{E}||_{2,1} + \lambda ||\mathbf{E}||_{G1}\\
& \nonumber \text{s.t.} \quad \mathbf{E} = [\mathbf{E}^{(1)}; \mathbf{E}^{(2)}; ...; \mathbf{E}^{(K)}], 
\\
& \nonumber ~~~~~~ \mathbf{P}^{(i)} = \mathbf{\hat{P}} + \mathbf{E}^{(i)}, 
\mathbf{\hat{P}} \leq 0, \mathbf{\hat{P}} \textbf{1} = \textbf{1}
\end{align}
\noindent where $rank(\mathbf{\hat{P}})$ is the rank of $\mathbf{\hat{P}}$. $\textbf{1}$ denotes the vector with all ones, $\beta$ and $\lambda$ are non-negative trade-off parameters. $\mathbf{E}$ is obtained by concatenation of error matrices of views vertically. To enforce $\mathbf{\hat{P}}$ to be a transition probability matrix, two constraints $\mathbf{\hat{P}} \geq 0$ and $\mathbf{\hat{P}} \textbf{1} = \textbf{1}$ have been considered.

\begin{figure}[t]
\centering
\includegraphics[width=0.45\textwidth,height=0.15\textheight]{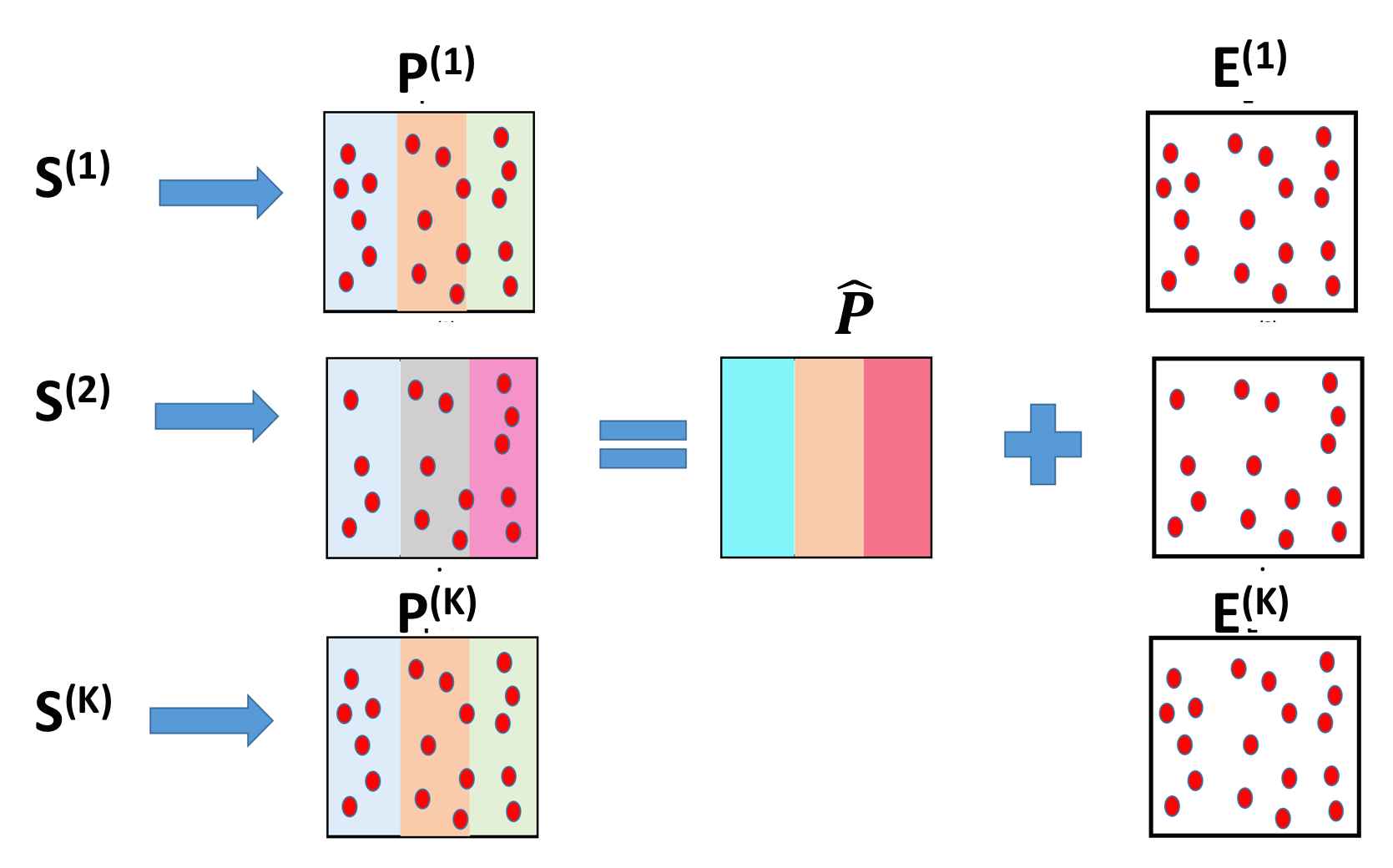}
\caption{Overview of the proposed EMVC method}
\label{fig:im}
\end{figure}

Since $rank(\mathbf{\hat{P}})$ is non-convex, the objective function in Eq. (5) is an instance of NP-hard problem. One natural way is to replace $rank(\mathbf{\hat{P}})$ with the trace norm $||\mathbf{\hat{P}}||_{*}$. The resulted objective function is as follows:
\begin{gather}
\label{for:5}
\min_{\mathbf{\hat{P}}, \mathbf{E}} ||\mathbf{\hat{P}}||_{*} + \beta ||\mathbf{E}||_{2,1} + \lambda ||\mathbf{E}||_{G1}\\
\nonumber \quad \text{s.t.} \quad \mathbf{E} = [\mathbf{E}^{(1)}; \mathbf{E}^{(2)}; ...; \mathbf{E}^{(K)}], \\
\nonumber \mathbf{P}^{(i)} = \mathbf{\hat{P}} + \mathbf{E}^{(i)}, \mathbf{\hat{P}} \leq 0, \mathbf{\hat{P}} \textbf{1} = \textbf{1}
\end{gather}
The trace norm is the convex envelope of the rank. Therefore, minimizing the trace norm of a matrix often applies the low-rank structure on that \cite{945730,srebro2004maximum}. Fig. \ref{fig:im} visualizes the proposed EMVC method. We first build initial transition probability matrices that might be erroneous. Then, we use decomposition via low-rankness and regularization to obtain a shared transition probability matrix.

\subsection{Optimization Procedure}
The objective function in Eq. (6) imposes a probabilistic simplex constraint on each of rows of $\mathbf{\hat{P}}$. We use an \textit{Augmented Lagrangian Multiplier} scheme \cite{LinChenMa2010} to solve the optimization problem. By introducing an auxiliary variable $\mathbf{Q}$, the objective function in Eq. (\ref{for:5}) can be stated equivalently as follows:
\begin{gather}
\label{for:6}
\min_{\mathbf{\hat{P}}, \mathbf{E}, \mathbf{Q}}  ||\mathbf{Q}||_{*} + \beta ||\mathbf{E}||_{2,1} + \lambda ||\mathbf{E}||_{G1}\\
\nonumber\quad \text{s.t.} \quad \mathbf{P}^{(i)} = \mathbf{\hat{P}} + \mathbf{E}^{(i)}, i = 1, ..., K. \\
\nonumber \mathbf{\hat{P}} \leq 0, \mathbf{\hat{P}} \textbf{1} = \textbf{1}, \mathbf{\hat{P}} = \mathbf{Q}
\end{gather}
\noindent where $\mathbf{Q} \in \mathbb{R}^{N \times N}$. The corresponding augmented Lagrangian function of Eq. (\ref{for:6}) is:
\begin{gather}
\label{for:7}
L(\mathbf{\hat{P}}, \mathbf{E}, \mathbf{Q}) = ||\mathbf{Q}||_{*} + \beta  ||\mathbf{E}||_{2,1} + \lambda  ||\mathbf{E}||_{G1} \nonumber \\
+ (\mu / 2)  ||\mathbf{\hat{P}} - \mathbf{Q}||^{2}_{F} 
+ \langle \mathbf{Z}, \mathbf{\hat{P}} - \mathbf{Q} \rangle \nonumber \\
+ (\mu/2) \sum_{i = 1}^{K} || \mathbf{\hat{P}} + \mathbf{E}^{(i)} - \mathbf{P}^{(i)}||^{2}_{F} \nonumber \\
 + \sum_{i = 1}^{K} \langle \mathbf{Y}^{(i)}, \mathbf{\hat{P}} + \mathbf{E}^{(i)} - \mathbf{P}^{(i)} \rangle
\end{gather}
\noindent where $\mathbf{Z} \in \mathbb{R}^{N \times N}$ and $\mathbf{Y}^{(i)} \in \mathbb{R}^{N \times N}$ are the Lagrange multipliers. $\mu$ is an adaptive penalty parameter which can be adjusted efficiently according to \cite{lin2011linearized}. 

The iterative algorithm for solving Eq. (8) is shown in Algorithm \ref{our_method}. We detail each step as follows.

\textbf{Solving $\mathbf{Q}$.} When other variables are fixed, the objective function w.r.t. $\mathbf{Q}$ can be stated as:
\begin{equation}
\min_{\mathbf{Q}} ||\mathbf{Q}||_{*} + (\mu/2) ||\mathbf{\hat{P}} - \mathbf{Q}||_{F}^{2} + \langle \mathbf{Z}, \mathbf{\hat{P}} - \mathbf{Q} \rangle
\label{for:8}
\end{equation}
The optimization problem in Eq. (\ref{for:8}) is equivalent to the following objective function:
\begin{equation}
\min_{\mathbf{Q}} ||\mathbf{Q}||_{*} + (\mu / 2) ||\mathbf{\hat{P}} - \mathbf{Q} + \mathbf{Z} / \mu||_{F}^{2}
\label{for:9}
\end{equation}
To solve Eq. (\ref{for:9}), we use the Singular Value Threshold (SVD) method \cite{doi:10.1137/080738970}. Let $\mathbf{U} \mathbf{\sum} \mathbf{V}^{T}$ denote the SVD form of $(\mathbf{\hat{P}} + \mathbf{Z}/\mu)$, then $\mathbf{Q}$ can be obtained using the following equation:
\begin{equation}
{\small
\mathbf{Q} = \mathbf{U} \mathbf{S}_{1/\mu}(\mathbf{\sum}) \mathbf{V}^{T}
}
\label{for:10}
\end{equation}
\noindent where $\mathbf{S_{\sigma}} = max(\mathbf{X} - \sigma, 0) + min(\mathbf{X} + \sigma, 0)$ is the shrinkage operator \cite{LinChenMa2010}.

\begin{algorithm}[t]
\caption{Error Robust Multi-View Clustering (EMVC)}
\label{our_method}
\begin{algorithmic}[1]
\REQUIRE $\lambda$, $\beta$, $\mathbf{P}^{(i)} \in \mathbb{R}^{N \times N} (i=1, 2, ..., K)$
\ENSURE Clustering results \\
 $\mathbf{\hat{P}} = 0$, $\mathbf{Q} = 0$, $\mathbf{Z} = 0$, $\mathbf{Y}^{(i)} = 0$, $\mathbf{E}^{(i)} = rand$, $\mu = 10^{-6}$, $\rho = 1.9$, $max_{\mu} = 10^{10}$, $\epsilon = 10^{-8}$
 
 \WHILE {not converge}
 \STATE $\mathbf{C} = (1/(K + 1)) (\mathbf{Q} - \mathbf{Z} /\mu + \sum_{i = 1}^{K} (\mathbf{P}^{(i)} - \mathbf{E}^{(i)} - \mathbf{Y}^{(i)} / \mu))$ 
 \FOR{i = 1, ..., $N$}
 \STATE Run Algorithm 4 using $\mathbf{c}^{i}$ as input to update $\mathbf{\hat{p}}^{i}$
 Where $c^{i}/\mathbf{\hat{p}}^{i}$ is the $i$-th row of $\mathbf{C}/\mathbf{\hat{P}}$, respectively.
 \ENDFOR
 \FOR{l = 1, ..., $N$}
 \STATE Update $\mathbf{e}_{l}$ by Eq. (\ref{for:E})
 \ENDFOR
 \STATE Update $\mathbf{Q}$ by Eq. (\ref{for:10})
 \STATE Update $\mathbf{Z}$ by Eq. (\ref{for:Z})
 \FOR{i = 1,..., $K$}
 \STATE Update $\mathbf{Y}^{(i)}$ by Eq. (\ref{for:Y})
 \ENDFOR
 \STATE $\mu \leftarrow min(\rho\mu, max_{\mu})$
\ENDWHILE
\STATE Apply Algorithm 2 on $\mathbf{\hat{P}}$ to get clustering results
\end{algorithmic}
\end{algorithm}

\textbf{Solving $\mathbf{E}$.} When other variables are fixed, the objective function w.r.t. $\mathbf{E}$ can be stated as:
\begin{align}
\label{for:11}
\min_{\mathbf{E}} \beta ||\mathbf{E}||_{2, 1} & + \lambda ||\mathbf{E}||_{G1} + 
 (\mu/2) \sum_{i = 1}^{K} ||\mathbf{\hat{P}} + \mathbf{E}^{(i)} - \mathbf{P}^{(i)}||_{F}^{2} \nonumber \\
& + \sum_{i = 1}^{K} \langle \mathbf{Y}^{(i)}, \mathbf{\hat{P}} + \mathbf{E}^{(i)} - \mathbf{P}^{(i)} \rangle
\end{align}
The optimization problem in Eq. (12) can be stated equivalently as follows:
\begin{gather}
\min_{\mathbf{E}} (\beta/\mu) ||\mathbf{E}||_{2,1} + (\lambda/\mu) ||\mathbf{E}||_{G1} +  (1/2) ||\mathbf{E} - \mathbf{B}||^{2}_{F}
\label{for:12}
\end{gather}
where $\mathbf{B}$ is constructed by vertically concatenating the matrices $\mathbf{P}^{(i)} - \mathbf{\hat{P}} - (1/\mu)\mathbf{Y}^{(i)}$ together along column.
Taking the derivative of Eq. (13) w.r.t. $\mathbf{E}$ and setting it to zero, we have the following result for $1 \leq l \leq N$:

\begin{gather}
(\beta/\mu) \mathbf{\hat{D}} \mathbf{e}_{l} + (\lambda/\mu) \mathbf{D}^{l} \mathbf{e}_{l} + (\mathbf{e}_{l} - \mathbf{p}_{l}^{(i)} + \mathbf{\hat{p}}_{l} - (1/\mu) \mathbf{y}_{l}^{(i)}) = 0
\label{for:13}
\end{gather}

\noindent where $\mathbf{e}_{l}$ denotes $\mathbf{E}(:,l)$ (i.e., $l$-th column of $\mathbf{E}$). Likewise, $\mathbf{p}_{l}^{(i)}$, $\mathbf{\hat{p}}_{l}$ and $\mathbf{y}_{l}^{(i)}$ indicate $l$-th column of $\mathbf{p}^{(i)}$, $\mathbf{\hat{p}}$ and $\mathbf{y}^{(i)}$, respectively.
$\mathbf{D}^{l}$ is a block diagonal matrix with the $j$-th diagonal block as $(1/(2||\mathbf{e}_{l}^{j}||_{2})) \times \mathbf{I}$, $\mathbf{I}$ is an identity matrix with size of $N$, $\mathbf{e}_{l}^{j}$ is the $j$-th segment of $\mathbf{e}_{l}$ and includes the representation errors in $j$-th view. $\mathbf{\hat{D}}$ is a diagonal matrix with the $i$-th diagonal element as $1/(2||\mathbf{e}^{i}||_{2})$ where $\mathbf{e}^{i}$ represents $i$-th row of $\mathbf{E}$. $\mathbf{e}_{l}$ can be obtained by:
\begin{equation}
\mathbf{e}_{l} = ((\beta/\mu) \mathbf{\hat{D}} + (\lambda/\mu) \mathbf{D}^{l} + \mathbf{I})^{-1} (\mathbf{P}^{(i)} - \mathbf{\hat{P}} + (1/\mu)\mathbf{Y}^{(i)})
\label{for:E}
\end{equation}

\textbf{Solving $\mathbf{\hat{P}}$.} Fixing other variables, we need to solve the following objective function to obtain $\mathbf{\hat{P}}$:
\begin{gather}
\min_{\mathbf{\hat{P}}} \sum_{i = 1}^{K} \langle \mathbf{Y}^{(i)}, \mathbf{\hat{P}} + \mathbf{E}^{(i)} - \mathbf{P}^{(i)} \rangle \\
 \nonumber + (\mu/2) \sum_{i = 1}^{K} ||\mathbf{\hat{P}} + \mathbf{E}^{(i)} - \mathbf{P}^{(i)}||^{2}_{F} \\
 \nonumber + \langle \mathbf{Z}, \mathbf{\hat{P}} - \mathbf{Q} \rangle  + (\mu/2) ||\mathbf{\hat{P}} - \mathbf{Q}||^{2}_{F} \quad \text{s.t.} \quad \mathbf{\hat{P}} \geq 0, \mathbf{\hat{P}}\textbf{1} = \textbf{1}
\label{for:14}
\end{gather}
The objective function in Eq. (16) can be converted to the following equivalent form:
\begin{gather}
\min_{\mathbf{\hat{P}}} (\mu/2) \sum_{i = 1}^{K} ||\mathbf{\hat{P}} + \mathbf{E}^{(i)} - \mathbf{P^{(i)}} + \mathbf{Y}^{(i)} / \mu||^{2}_{F} \\
\nonumber  + (\mu/2) ||\mathbf{\hat{P}} - \mathbf{Q} + \mathbf{Z} /\mu||^{2}_{F}
\quad \text{s.t.} \quad \mathbf{\hat{P}} \geq 0, \mathbf{\hat{P}}\textbf{1} = \textbf{1}
\label{for:15}
\end{gather}
For ease of presentation, we define a new variable $\mathbf{C}$ as follows: 
\begin{equation}
\mathbf{C} = (1/(K + 1)) (\mathbf{Q} - \mathbf{Z} /\mu + \sum_{i = 1}^{K} (\mathbf{P}^{(i)} - \mathbf{E}^{(i)} - \mathbf{Y}^{(i)} / \mu))
\end{equation}
Then the objective function in Eq. (17) can be converted into the following equivalent form:
\begin{equation}
\min_{\mathbf{\hat{P}}} (1/2)||\mathbf{\hat{P}} - \mathbf{C}||^{2}_{F} \hspace{0.3cm} s.t. \quad \mathbf{\hat{P}} \geq 0, \mathbf{\hat{P}}\textbf{1} =\textbf{1}
\end{equation}
Eq.~(19) can be further rewritten as:
\begin{gather}
\min_{\mathbf{\hat{p}}^{1}, ..., \mathbf{\hat{p}}^{N}} (1/2)\sum_{i = 1}^{N} ||\mathbf{\hat{p}}^{i} - \mathbf{c}^{i}||_{F}^{2}
\hspace{0.2cm} \text{s.t.} \quad \sum_{i = 1}^{K} \hat{p}_{i,j} = 1, \hat{p}_{i,j} \geq 0
\label{for:17}
\end{gather}
\noindent where $\mathbf{c}^{i}$ indicates the $i$-th row of matrix $\mathbf{C}$ and $\mathbf{\hat{p}}^{i}$ denotes the $i$-th row of matrix $\mathbf{\hat{P}}$, respectively. The optimization problem in Eq. (20) has $N$ independent subproblems. Each problem is a proximal operator problem with a probabilistic simplex constraint that can be efficiently solved by the projection algorithm \cite{Duchi:2008:EPL:1390156.1390191}. The algorithm for this optimization procedure is shown in Algorithm \ref{P}.

\textbf{Solving $\mathbf{Z}$.} The Lagrangian multiplier $\mathbf{Z}$ can be obtained using the following update:
\begin{equation}
\mathbf{Z} \leftarrow \mathbf{Z} + \mu (\mathbf{\hat{P}} - \mathbf{Q})
\label{for:Z}
\end{equation}

\textbf{Solving $\mathbf{Y}^{(i)}$.} The Lagrangian multiplier $\mathbf{Y}^{(i)}$ can be obtained using the following update:
\begin{equation}
\mathbf{Y}^{(i)} \leftarrow \mathbf{Y}^{(i)} + \mu (\mathbf{\hat{P}} + \mathbf{E}^{(i)} - \mathbf{P}^{(i)})
\label{for:Y}
\end{equation}

\begin{algorithm}[t]
\caption{Proximal Operator with Simplex Constraint}
\label{P}
\begin{algorithmic}[1]
\REQUIRE $\mathbf{c}^{i} \in \mathbb{R}^{N}$
\ENSURE $\mathbf{\hat{p}}^{i} \in \mathbb{R}^{N}$ \\
 \STATE $\mathbf{u} = Sort(\mathbf{c}^{i}, 'descend')$
 \STATE $\hat{j} = max\{j: 1-\sum_{r = 1}^{j} (u_{r} - u_{j}) \geq 0\}$
 \STATE $\sigma = (1/\hat{j})(\sum_{i=1}^{\hat{j}} u_{i} - 1)$
 \FOR{j = 1, ..., N}
 \STATE $\hat{p}_{i,j} = max(c_{i,j} - \sigma, 0)$
 \ENDFOR
\end{algorithmic}
\end{algorithm}

\subsection{Computational and Convergence analysis}
In Algorithm 3, Lines 2-5 update $\mathbf{\hat{P}}$ with quadratic complexity. Lines 6-8 update matrix $\mathbf{E}$. Instead of computing the matrix inverse with cubic complexity, we can solve a system of linear equations which have quadratic complexity in order to obtain $e_{l}$. If sufficient computational resources are available, each $e_{l}$ ($1 \leq l \leq N$) can be computed in parallel with efficiency. Updating $\mathbf{Q}$ in line 9 requires solving an SVD problem. This part can be computed with cubic complexity. Lagrangian Multipliers can be updated with quadratic complexity. Line 16 applies spectral clustering via Markov chains (i.e., Algorithm 2) on the shared transition probability matrix. This step can be done with cubic complexity. When sufficient computational resources are available and parallel computing is implemented, both SVD and linear equations can be solved efficiently.

For convergence analysis, the following theorem guarantees the convergence of Algorithm 3.

\textbf{Theorem.} Algorithm 3 decreases the objective value of Eq. (6) in each iteration.

\textbf{Proof.} To obtain $\mathbf{\hat{P}}_{t+1}$ (i.e., $\mathbf{\hat{P}}$ in ($t+1$)-th iteration), according to Algorithm 3, we know that
\begin{gather}
 \nonumber \mathbf{\hat{P}}_{t+1} = \min_{\mathbf{\hat{P}}} ||\mathbf{\hat{P}}||_{*} + \lambda \sum_{i = 1}^{N} \mathbf{D}_{t+1}^{i} ||(\mathbf{E}_{t})_{i}||_{2}^{2} \\
 + \beta Tr(\mathbf{E}_{t}^{T} \mathbf{\hat{D}}_{t+1} \mathbf{E}_{t}) 
\label{p23}
\end{gather}
\noindent where $\mathbf{E}_{t}$ represents $\mathbf{E}$ at $t$-th iteration and $(\mathbf{E}_{t})_{i}$ indicates $i$-th column of $\mathbf{E}_{t}$. According to Algorithm 3, to obtain $\mathbf{E}_{t+1}$, the following problem must be solved:
\begin{gather}
\mathbf{\hat{E}}_{t+1} = \min_{\mathbf{E}} ||\mathbf{\hat{P}}_{t+1}||_{*} + \lambda \sum_{i = 1}^{N} \mathbf{D}_{t+1}^{i} ||(\mathbf{E}_{t})_{i}||_{2}^{2} \nonumber  \\
+ \beta Tr(\mathbf{E}^{T} \mathbf{\hat{D}}_{t+1} \mathbf{E}) 
\label{p24}
\end{gather}
Considering Eq. (23) and Eq. (24), we have the following:
\begin{gather}
\nonumber ||\mathbf{\hat{P}}_{t+1}||_{*} + \lambda \sum_{i = 1}^{N} \mathbf{D}_{t+1}^{i} ||(\mathbf{E}_{t})_{i}||_{2}^{2} + \beta Tr(\mathbf{E}_{t+1}^{T} \mathbf{\hat{D}}_{t+1} \mathbf{E}_{t+1})\\
 \leq ||\mathbf{\hat{P}}_{t}||_{*} + \lambda \sum_{i = 1}^{N} \mathbf{D}_{t+1}^{i} ||(\mathbf{E}_{t})_{i}||_{2}^{2} + \beta Tr(\mathbf{E}_{t}^{T} \mathbf{\hat{D}}_{t+1} \mathbf{E}_{t})
\label{eq:eqj}
\end{gather}
Substituting $\mathbf{D}$ and $\mathbf{\hat{D}}$ by their definitions results in the following:
\begin{gather}
 ||\mathbf{\hat{P}}_{t}||_{*} + \lambda \sum_{i = 1}^{N} \sum_{j = 1}^{K} ||(\mathbf{E}_{t})^{j}_{i}||_{2}^{2}/(2||((\mathbf{E_{t}})_{i}^{j}||_{2}) \\
 \nonumber + \beta \sum_{i = 1}^{N \times K} ||(\mathbf{E}_{t})^{i}||^{2}_{2} /(2||(\mathbf{E}_{t})^{i}||_{2})
\end{gather}
\noindent where $(\mathbf{E}_{t})_{i}^{j}$ denotes the segment $((j-1)*N+1:j*N)$ of $i^\text{th}$ row of $\mathbf{E}_{t}$ and $(\mathbf{E}_{t})^{i}$ indicates $i^\text{th}$ row of $\mathbf{E}_{t}$. We can derive the following because if we define $f(x) = x - x^{2}/2\alpha$, then $f(x) \leq f(\alpha)$:
\begin{gather}
 \nonumber \sum_{j = 1}^{K} ||(\mathbf{E}_{t+1})^{j}_{i}||_{2} - \sum_{j = 1}^{K} ||(\mathbf{E}_{t+1})_{i}^{j}||^{2}_{2} /(2||(\mathbf{E}_{t})_{i}^{j}||_{2}) \\
 \leq \sum_{j = 1}^{K} ||(\mathbf{E}_{t})_{i}^{j}||_{2} - \sum_{j = 1}^{K} ||(\mathbf{E}_{t})_{i}^{j}||^{2}_{2} /(2||(\mathbf{E}_{t})_{i}^{j}||_{2})
\end{gather}
and
{\small
\begin{gather}
\nonumber \sum_{i = 1}^{K \times N} ||(\mathbf{E}_{t+1})^{i}||_{2} - \sum_{i = 1}^{K \times N} ||(\mathbf{E}_{t+1})^{i}||^{2}_{2} /(2||(\mathbf{E}_{t})^{i}||_{2}) \\
  \leq \sum_{i = 1}^{K \times N} ||(\mathbf{E}_{t})^{i}||_{2} - \sum_{i = 1}^{K \times N} ||(\mathbf{E}_{t})^{i}||^{2}_{2} /(2||(\mathbf{E}_{t})^{i}||_{2})
\end{gather}
}
If we add all Eq. (25-28) on both sides, we obtain the inequality that objective value at iteration $t+1$ is less than objective value at iteration $t$.
Therefore, we can conclude that the objective value decreases in each iteration and Algorithm 3 converges.
\section{Experimental Evaluation}
To empirically evaluate performance of the proposed EMVC method, we conduct extensive experiments on synthetic and publicly available real-world multi-view datasets and compare with six state-of-the-art methods: 
(1) \textbf{Best Single View (BSV)} performs standard \textit{k-means} on the most informative view.
(2) \textbf{Feature Concatenation (Feat. Concat.)} concatenates features of all views and then runs standard \textit{k}-means clustering on the concatenated feature representations.
(3) \textbf{Kernel Addition} constructs kernel matrix for each individual view and then obtains the average of these matrices to get a single kernel matrix for spectral clustering.
(4) \textbf{Co-regularized Spectral Clustering (Co-Reg)} performs centroid based and pairwise co-regularized spectral clustering via Gaussian kernel \cite{Kumar:2011:CMS:2986459.2986617}. The co-regularization parameter $\lambda$ is tuned by searching a range of $\{0.01, 0.02, 0.03, 0.04, 0.05\}$ as suggested by the authors.
(5) \textbf{Robust Multi-View Spectral Clustering via Low-Rank and Sparse Decomposition (RMSC)} uses low-rank decomposition and $\ell_1$ norm \cite{Xia:2014:RMS:2892753.2892850}. The regularization parameter $\lambda$ is tuned by searching the range \{$10^{-3}$, $10^{-2}$, ..., $10^{2}$, $10^{3}$\} as suggested by the authors (we keep $\lambda$ the same for all views). The parameter $\sigma^{2}$ is set to the median of all Euclidean distances over all pairs of data points for each individual view as suggested by the authors. 
(6) \textbf{Parameter-Free Auto-Weighted Multiple Graph Learning (AMGL)} finds cluster indicator matrix over all views by applying normalized cut algorithms on the graphs of views \cite{Nie:2016:PAM:3060832.3060884}. 

We implement four versions of the proposed EMVC method to investigate the effectiveness of its component terms in multi-view learning: EMVC by only using the first term in Eq. (6) ``EMVC(*)", EMVC by using trace norm and only imposing the group $\ell_{1}$ norm ``EMVC($g_{1}$)", EMVC by using trace norm and only imposing $\ell_{2,1}$ norm ``EMVC($\ell_{2,1}$)" and the full version of EMVC based on Eq. (6). We apply grid search to identify optimal values for each regularization hyperparameter from $\{10^{-9}, 10^{-8}, ..., 10^{8}, 10^{9}\}$. The standard deviation is set to the median of all Euclidean distances over all pairs of data points for each individual view.

We use different evaluation metrics including \textbf{F-Score}, \textbf{Precision}, \textbf{Recall}, \textbf{Normalized Mutual Information (NMI)}, \textbf{Entropy}, \textbf{Accuracy}, \textbf{Adjusted Rand-Index (AR)} for the purpose of comprehensive evaluation \cite{Kumar:2011:CMS:2986459.2986617,7298657}. 
All of these measure except for Entropy are positive measures, which indicates that larger values stand for better performance. For Entropy, smaller values indicate better performance. Different measurements reveal different properties. Thus, we can obtain comprehensive view from the results.  

Each experiment is repeated for five times, and the mean and standard deviation of each metric in each dataset are reported. We then use \textit{k}-means to obtain final clustering solution. 
Since \textit{k}-means is sensitive to initial seed selection, we run \textit{k}-means 20 times on each dataset. 

\vspace{-0.1cm}



\subsection{Experiments on Real-World Datasets}
We conduct experiments on the following publicly available real-world datasets. Statistics of the real-world datasets are summarized in Table \ref{tab:dataset} (Max \# features indicates maximum number of features over all views of the dataset).

\begin{table}[ht]
\centering
\caption{Statistics of the real-world multi-view datasets}
\label{tab:dataset}
\begin{tabular}{c|c|c|c|c}
\hline
\textbf{Dataset} & \textbf{\# data points} & \textbf{Max \# features} & \textbf{\# views} & \textbf{\# clusters}\\
\hline


webKB    & 1051 & 3000 & 2 & 2 \\           

FOX      & 1523 & 2711 & 2 & 4 \\

CNN      & 2107 & 3695 & 2 & 7 \\   

Citeseer & 3312 & 6654 & 2 & 6 \\

CCV & 9317 & 5000 & 3 & 20 \\
\hline

\end{tabular}
\end{table}

\begin{table*}[ht]
\centering
\caption{Comparison results on the real-world datasets - part 1(mean (standard deviation))}
\label{tab:result1}
\begin{tabular}{c|c|c|c|c|c|c|c|c}
\hline
\textbf{Dataset} & \textbf{Method} & \textbf{F-Score} $\uparrow$ & \textbf{Precision} $\uparrow$ & \textbf{Recall} $\uparrow$ & \textbf{NMI} $\uparrow$ & \textbf{Entropy} $\downarrow$ & \textbf{Accuracy} $\uparrow$ & \textbf{AR} $\uparrow$\\
\hline
WebKB    & BSV      & 0.889(0.000) & 0.889(0.000) & 0.889(0.000) & 0.532(0.000) & 0.406(0.000) & 0.913(0.000) & 0.618(0.000)\\
         & Feat. Concat. & 0.947(0.000) & 0.947(0.000) & 0.947(0.000) & 0.717(0.000) & 0.214(0.000) & 0.963(0.000) & 0.845(0.000)\\
         & Kernel Addition       & 0.946(0.000) & 0.946(0.000) & 0.946(0.000 & 0.717(0.000) & 0.214(0.000) & 0.963(0.000) & 0.845(0.000)\\
         & Co-Reg       & 0.949(0.000) & 0.949(0.000) & 0.949(0.000) & 0.733(0.000) & 0.195(0.000) & 0.965(0.000)  & 0.853(0.000)\\
         & AMGL                  & 0.794(0.000) & 0.794(0.000) & 0.794(0.000) & 0.015(0.000) & 0.752(0.000) & 0.783(0.000) & 0.013(0.000)\\
         & RMSC                  & 0.956(0.000) & 0.956(0.000) & 0.956(0.000) & 0.758(0.000) & 0.189(0.000) & 0.970(0.000) & 0.871(0.000)\\
         & EMVC (*)              & 0.568(0.000) & 0.568(0.000) & 0.568(0.000) & 0.000(0.000) & 0.757(0.000) & 0.511(0.000) & 0.000(0.000) \\
         & EMVC ($g_{1}$)        & 0.949(0.000) & 0.949(0.000) & 0.949(0.000) & 0.731(0.000) & 0.199(0.000) & 0.965(0.000) & 0.853(0.000) \\
         & EMVC ($\ell_{2,1}$)    & 0.954(0.000) & 0.954(0.000) & 0.954(0.000) & 0.759(0.000) & 0.175(0.000) & 0.969(0.000) & 0.870(0.000) \\
         & EMVC                 & \textbf{0.959}(0.000) & \textbf{0.959}(0.000) & \textbf{0.959}(0.000) & \textbf{0.776}(0.000) & \textbf{0.164}(0.000) & \textbf{0.972}(0.000) & \textbf{0.883}(0.000)\\
\hline
FOX      & BSV      & 0.718(0.000) & 0.718(0.000) & 0.718(0.000) & 0.672(0.000) & 0.626(0.000) & 0.758(0.000)  & 0.599(0.000)\\
         & Feat. Concat. & 0.314(0.000) & 0.314(0.000) & 0.314(0.000) & 0.041(0.000) & 1.787(0.000) & 0.356(0.000) & 0.050(0.000) \\
         & Kernel Addition       & 0.358(0.000) & 0.358(0.000) & 0.358(0.000) & 0.103(0.000) & 1.669(0.000) & 0.460(0.000) & 0.113(0.000) \\
         & Co-Reg       & 0.477(0.006) & 0.477(0.006) & 0.477(0.006) & 0.242(0.002) & 1.410(0.002) & 0.547(0.000) & 0.262(0.000)\\
         & AMGL                  & 0.456(0.000) & 0.456(0.000) & 0.456(0.000) & 0.010(0.000) & 1.857(0.000) & 0.419(0.000) & 0.001(0.000)\\
         & RMSC                  & 0.364(0.005) & 0.364(0.005) & 0.364(0.005) & 0.141(0.000) & 1.593(0.001) & 0.401(0.001) & 0.127(0.000)\\
         & EMVC (*)  & 0.270(0.009) & 0.270(0.009) & 0.270(0.009) & 0.002(0.002) & 1.862(0.002) & 0.267(0.003) & 0.000(0.000)\\
         & EMVC ($g_{1}$)   & 0.761(0.005) & 0.761(0.005) & 0.761(0.005) & 0.691(0.003) & 0.565(0.002) & 0.818(0.004) & 0.664(0.002)\\
         & EMVC ($\ell_{2,1}$) & 0.761(0.004) & 0.761(0.004) & 0.761(0.004) & 0.691(0.007) & 0.565(0.004) & 0.818(0.003) & 0.664(0.005)\\
         & EMVC                 & \textbf{0.761}(0.010) & \textbf{0.761}(0.010) & \textbf{0.761}(0.010) & \textbf{0.691}(0.004) & \textbf{0.565}(0.003) & \textbf{0.818}(0.003) & \textbf{0.664}(0.002)\\
\hline
CNN      & BSV      & 0.388(0.001) & 0.388(0.001) & 0.388(0.001) & 0.405(0.008) & 1.736(0.012) & 0.486(0.007) & 0.228(0.008)\\
         & Feat. Concat. & 0.171(0.000) & 0.171(0.000) & 0.171(0.000) & 0.037(0.000) & 2.621(0.001) & 0.219(0.001) & 0.023(0.000)\\
         & Kernel Addition       & 0.175(0.000) & 0.175(0.000) & 0.175(0.000) & 0.046(0.000) & 2.597(0.002) & 0.233(0.000) & 0.026(0.000)\\
         & Co-Reg       & 0.200(0.005) & 0.200(0.005) & 0.200(0.005) & 0.076(0.002) & 2.513(0.003) & 0.276(0.002) & 0.056(0.004)\\
         & AMGL                  & 0.250(0.002) & 0.250(0.002) & 0.250(0.002) & 0.031(0.001) & 2.667(0.003) & 0.239(0.004) & 0.000(0.001)  \\
         & RMSC                  & 0.219(0.010) & 0.219(0.010) & 0.219(0.010) & 0.122(0.000) & 2.388(0.001) & 0.300(0.000) & 0.078(0.000) \\
         & EMVC (*) & 0.149(0.003) & 0.149(0.003) & 0.149(0.003) & 0.003(0.004) & 2.716(0.004) & 0.165(0.002) & 0.000(0.000) \\
         & EMVC ($g_{1}$)   & 0.557(0.004) & 0.557(0.004) & 0.557(0.004) & 0.536(0.002) & 1.279(0.005) & 0.655(0.005) & 0.472(0.000)\\
         & EMVC ($\ell_{2,1}$)      & 0.558(0.004) & 0.558(0.004) & 0.558(0.004) & 0.536(0.003) & 1.281(0.004) & 0.656(0.001) & 0.472(0.001)\\
         & EMVC  & \textbf{0.560}(0.013) & \textbf{0.560}(0.013) & \textbf{0.560}(0.013) & \textbf{0.542}(0.005) & \textbf{1.264}(0.003) & \textbf{0.657}(0.002) & \textbf{0.474}(0.002)\\
\hline
\end{tabular}
\label{table-result1}
\end{table*}




\textbf{WebKB}\footnote{\url{http://www.cs.cmu.edu/afs/cs/project/theo-20/www/data/}}: This dataset contains webpages collected from Texas, Cornell, Washington and Wisconsin universities. Each webpage is described by the content view and link view.

\textbf{FOX}\footnote{\url{https://sites.google.com/site/qianmingjie/home/datasets/}}: The dataset is crawled from FOX web news. Each instance is represented in two views: the text view and the image view. Titles, abstracts, and text body contents are extracted as the text view data, and the image included in the article is stored as the image view data. 

\textbf{CNN}\footnote{\url{https://sites.google.com/site/qianmingjie/home/datasets/}}: This dataset is crawled from CNN web news. For this dataset, titles, abstracts, and text body contents are extracted as the text view data. Also, the image included in the article is stored as the image view data. 

\textbf{Citeseer}\footnote{\url{http://linqs.cs.umd.edu/projects//projects/lbc/index.html}}: It contains a selection of the Citeseer dataset. The papers were selected in a way that in the final corpus every paper cites or is cited by at least one other paper. The text view consists of title and abstract of a paper; the link view contains inbound and outbound references.

\textbf{CCV}\footnote{\url{http://www.ee.columbia.edu/ln/dvmm/CCV/}}: This high rank dataset contains 9317 videos over 20 semantic categories. Two views contains visual features, while the third view consists of audio features. 

\begin{table*}[t]
\centering
\caption{Comparison results on the real-world datasets - part 2(mean (standard deviation))}
\label{tab:result1}
\begin{tabular}{c|c|c|c|c|c|c|c|c}
\hline
\textbf{Dataset} & \textbf{Method} & \textbf{F-Score} $\uparrow$ & \textbf{Precision} $\uparrow$ & \textbf{Recall} $\uparrow$ & \textbf{NMI} $\uparrow$ & \textbf{Entropy} $\downarrow$ & \textbf{Accuracy} $\uparrow$ & \textbf{AR} $\uparrow$\\
\hline
Citeseer & BSV      & 0.322(0.000) & 0.322(0.000) & 0.322(0.000) & 0.199(0.000) & 2.013(0.000) & 0.443(0.000) & 0.180(0.000)\\
         & Feat. Concat. & 0.326(0.001) & 0.326(0.001) & 0.326(0.001) & 0.204(0.002) & 2.001(0.001) & 0.452(0.001) & 0.185(0.000)\\
         & Kernel Addition       & 0.346(0.002) & 0.346(0.002) & 0.346(0.002) & 0.232(0.003) & 1.943(0.002) & 0.456(0.001) & 0.200(0.001)\\
         & Co-Reg       & 0.356(0.009) & 0.356(0.009) & 0.356(0.009) & 0.174(0.010) & 2.088(0.005) & 0.378(0.003) & 0.123(0.003) \\
         & AMGL                  & 0.303(0.010) & 0.303(0.010) & 0.303(0.010) & 0.005(0.009) & 2.517(0.007) & 0.213(0.002) & 0.000(0.001) \\
         & RMSC                  & 0.271(0.011) & 0.271(0.011) & 0.271(0.011) & 0.154(0.005) & 2.139(0.009) & 0.365(0.002) &  0.105(0.001)\\
         & EMVC (*) & 0.172(0.020) & 0.172(0.020) & 0.172(0.020) & 0.001(0.011) & 2.519(0.009) & 0.183(0.003) &  0.000(0.002)\\
         & EMVC ($g_{1}$)   & 0.386(0.006) & 0.386(0.006) & 0.386(0.006) & 0.283(0.007) & 1.800(0.005) & 0.532(0.004) & 0.251(0.002)\\
         & EMVC ($\ell_{2,1}$)      & 0.388(0.007)     & 0.388(0.007)     & 0.388(0.007)     & 0.284(0.008) & 1.802(0.004)     & 0.535(0.003)      & 0.254(0.002)\\
         & EMVC            & \textbf{0.390}(0.007) & \textbf{0.390}(0.007) & \textbf{0.390}(0.007) & \textbf{0.286}(0.011) & \textbf{1.803}(0.002) & \textbf{0.537}(0.004) & \textbf{0.256}(0.002)\\
\hline
CCV      & BSV           & 0.119(0.001) & 0.119(0.001) & 0.119(0.001) & 0.177(0.001) & 3.466(0.003) & 0.181(0.006)  & 0.069(0.002)\\
         & Feat. Concat. & 0.096(0.001) & 0.096(0.001) & 0.096(0.001) & 0.119(0.001) & 3.739(0.010) & 0.170(0.002) & 0.023(0.001)\\
         & Kernel Addition       & 0.124(0.002) & 0.124(0.002) & 0.124(0.002) & 0.171(0.001) & 3.496(0.005) & 0.189(0.009)  & 0.072(0.002)\\
         & Co-Reg        & 0.119(0.009) & 0.119(0.009) & 0.119(0.009) & 0.176(0.001) & 3.473(0.075) & 0.180(0.010) & 0.068(0.040)\\
         & AMGL                   & 0.080(0.010) & 0.080(0.010) & 0.080(0.010) & 0.089(0.001) & 3.901(0.009) & 0.165(0.006)  & 0.019(0.002)\\
         & RMSC                  & 0.130(0.005) & 0.130(0.005) & 0.130(0.005) & 0.203(0.002) & 3.225(0.020) & 0.196(0.005) & 0.082(0.005)\\
         & EMVC (*) & 0.070(0.005) & 0.070(0.005) & 0.070(0.005) & 0.085(0.006) & 4.001(0.004) & 0.152(0.009)  & 0.012(0.000)\\
         & EMVC ($g_{1}$)   & 0.131(0.004) & 0.131(0.004) & 0.131(0.004) & 0.210(0.007) & 3.100(0.003) & 0.198(0.004)  & 0.090(0.004)\\
         & EMVC ($\ell_{2,1}$)      & 0.131(0.004) & 0.131(0.004) & 0.131(0.004) & 0.211(0.006) & 3.090(0.004) & 0.198(0.005)  & 0.090(0.002)\\
         & EMVC            & \textbf{0.141}(0.009) & \textbf{0.141}(0.009) & \textbf{0.141}(0.009) & \textbf{0.300}(0.009) & \textbf{2.987}(0.002) & \textbf{0.203}(0.008)  & \textbf{0.091}(0.004)\\
\hline
\end{tabular}
\label{table-result2}
\end{table*}

\begin{table*}[t]
\centering
\caption{Comparison results on the synthetic dataset (mean (standard deviation))}
\label{tab:result1}
\begin{tabular}{c|c|c|c|c|c|c|c}
\hline
\textbf{Method} & \textbf{F-Score} $\uparrow$ & \textbf{Precision} $\uparrow$ & \textbf{Recall} $\uparrow$ & \textbf{NMI} $\uparrow$ & \textbf{Entropy} $\downarrow$ & \textbf{Accuracy} $\uparrow$  & \textbf{AR} $\uparrow$\\
\hline
         BSV      & 0.655(0.000) & 0.655(0.000) & 0.655(0.000) & 0.246(0.000) & 0.758(0.000) & 0.771(0.000) & 0.293(0.000)\\
         Feat. Concat. & 0.748(0.000) & 0.748(0.000) & 0.748(0.000) & 0.424(0.000) & 0.581(0.000) & 0.849(0.000) &  0.486(0.000)\\
         Kernel Addition       & 0.760(0.000) & 0.760(0.000) & 0.760(0.000) & 0.439(0.000) & 0.564(0.000) & 0.859(0.000) & 0.515(0.00)0\\
         Co-Reg       & 0.750(0.000) & 0.750(0.000) & 0.750(0.000) & 0.437(0.000) & 0.569(0.000) & 0.850(0.000) &  0.489(0.000) \\
         AMGL                  & 0.579(0.000) & 0.579(0.000) & 0.579(0.000) & 0.116(0.003) & 0.883(0.003) & 0.696(0.002) &  0.153(0.004)\\
         RMSC                  & 0.736(0.000)  & 0.736(0.000)  & 0.736(0.000) & 0.375(0.000) & 0.624(0.000) & 0.844(0.000) & 0.472(0.000)\\
         EMVC (*)  & 0.499(0.000) & 0.499(0.000) & 0.499(0.000) & 0.000(0.000) & 1.000(0.000) & 0.501(0.000) & 0.000(0.000)\\
         EMVC ($g_{1}$)   & 0.730(0.000) & 0.730(0.000) & 0.730(0.000) & 0.366(0.000) & 0.634(0.000) & 0.840(0.000) & 0.461(0.000)\\
         EMVC ($\ell_{2,1}$) & 0.730(0.000) & 0.730(0.000) & 0.730(0.000) & 0.366(0.000) & 0.634(0.000) & 0.840(0.000) & 0.461(0.000)\\
         EMVC & \textbf{0.762(0.000)} & \textbf{0.762(0.000)} & \textbf{0.762(0.000)} & \textbf{0.449(0.000)} & \textbf{0.555(0.000)} & \textbf{0.860(0.000)} & \textbf{0.517(0.000)}\\
\hline
\end{tabular}
\label{tab:syn1}
\end{table*}




Tables \ref{table-result1} and \ref{table-result2} report the performance comparison on the real-world datasets. 
From these tables, we have several observations. First, EMVC is always better than the baselines by a large margin. Specifically, compared with RMSC, even EMVC($g_1$) can do a lot better in some of the datasets like FOX, CNN and Citeseer. This observation is consistent with our analysis in that group $\ell_1$ achieves better performance than $\ell_1$.
Second, the full version of EMVC is superior to all its three degenerative versions. This validates the correctness of our objective function and demonstrates the importance of having an all-encompassing approach.

\subsection{Experiments on Synthetic Noisy Dataset}
Using similar settings in \cite{Kumar:2011:CMS:2986459.2986617}, the synthetic dataset consists of two views and the data in each view is partitioned into two clusters. Eq. (29) shows cluster means and covariances for each view.
In each view, the two clusters overlap, which is the source of noise in the transition probabilities of each view. 
First, we choose the cluster that each sample belongs to, and then produce the views from a mixture of two bivariate Gaussian distributions. For each view, we sample 500 data points from each of the clusters. 

\begin{equation}
\nonumber \mathbf{\mu}^{(1)}_{1} = (1, 1), \mathbf{\mu}^{(1)}_{2} = (2, 2), 
\mathbf{\mu}^{(2)}_{1} = (2, 2), \mathbf{\mu}^{(2)}_{2} = (1,1)
\end{equation}
\vspace{-0.6cm}
\begin{align}
{\sum}_{*} =
\left(
  \begin{tabular}{cc}
  1   & 0.5\\
  0.5 & 1.5
  \end{tabular}
\right), 
{\sum}_{**} =
\left(
  \begin{tabular}{cc}
  0.3 & 0\\
  0   & 0.6
  \end{tabular}
\right)\nonumber \\
\label{synth:1}
\end{align}
\noindent where $\mathbf{\mu}^{(i)}_{j}$ denotes cluster means for cluster $j$ in view $i$. ${\sum}_{*}$ is covariance for first and second clusters in first and second views, respectively. ${\sum}_{**}$ indicates covariance for second and first clusters in first and second views, respectively. 
Table \ref{tab:syn1} presents the comparison results on this dataset. With this type of noise, the proposed EMVC method shows superior clustering performance over all the baselines. 


 \begin{figure*}[ht]
        \centering
        \subfloat[NRC-FOX]{
                \centering
                \includegraphics[width=0.25\linewidth,height=3.4cm]{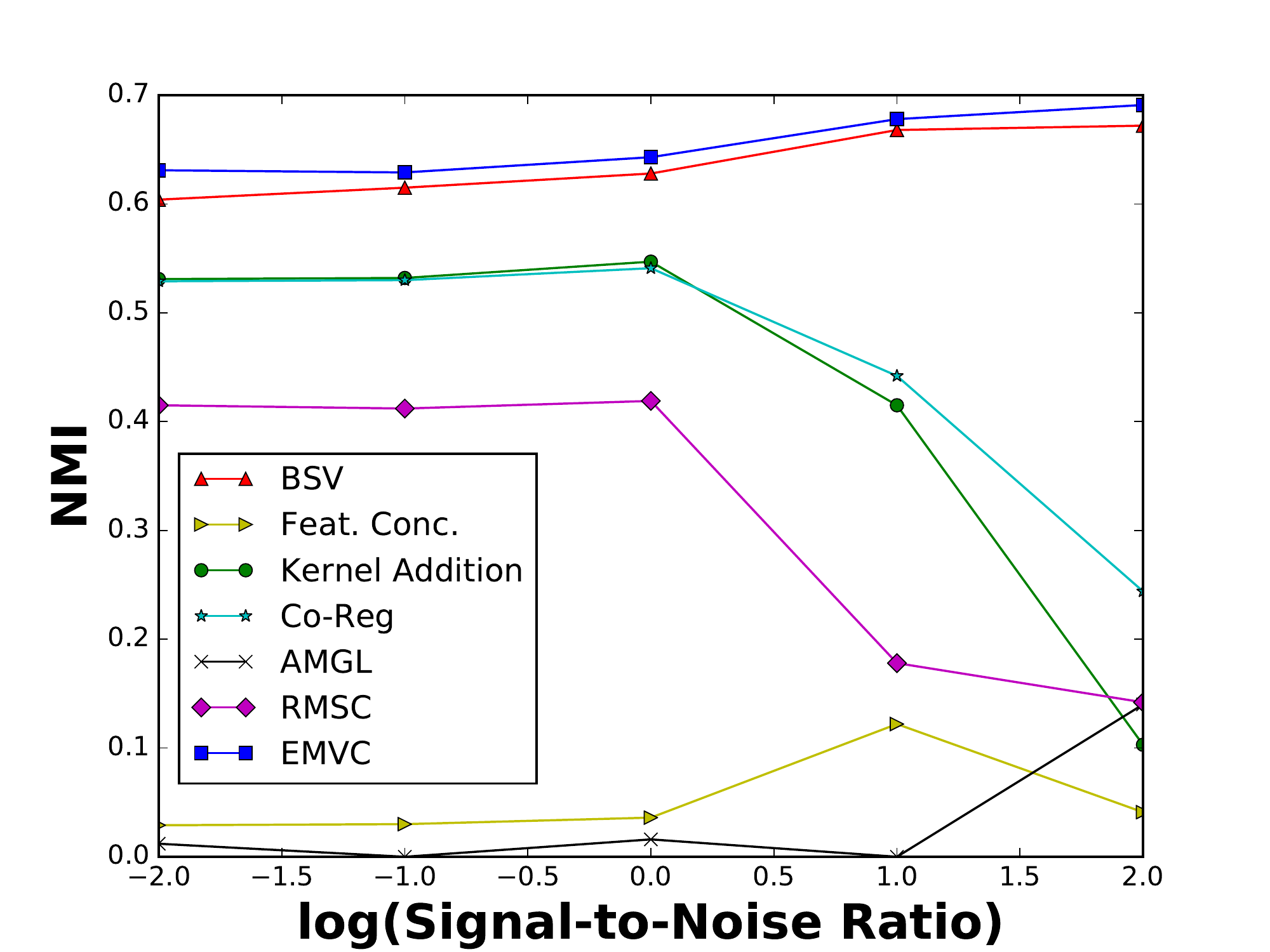}
                \label{fig:FOX-nmi}\hfill}
        \subfloat[NRC-CNN]{
                \centering
                \includegraphics[width=0.25\linewidth,height=3.4cm]{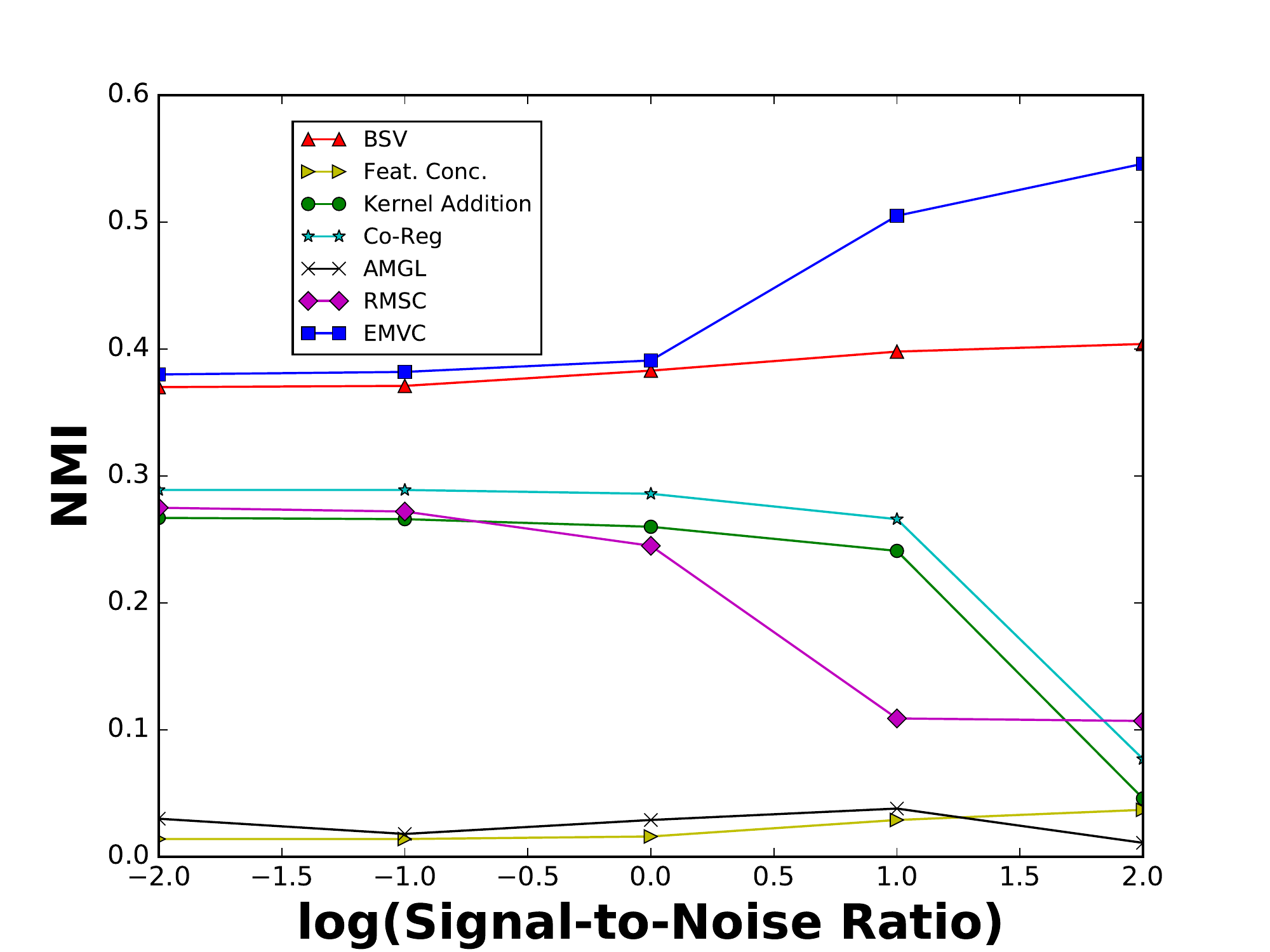}
                \label{fig:CNN-nmi}
        }
        \subfloat[SSC-FOX]{
                \centering
                \includegraphics[width=0.25\linewidth,height=3.4cm]{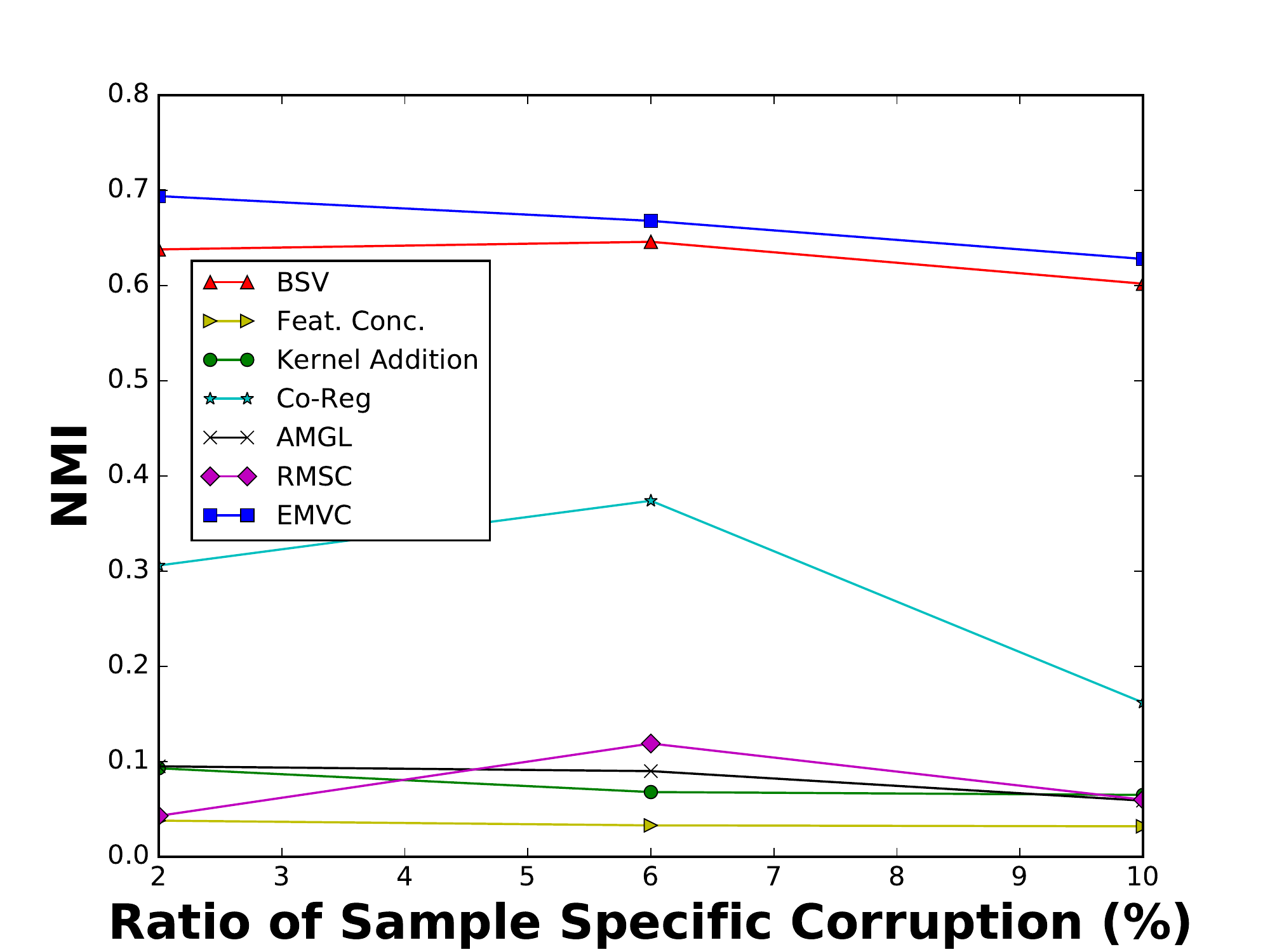}
                \label{fig:FOX-nmi}\hfill}
        \subfloat[SSC-CNN]{
                \centering
                \includegraphics[width=0.25\linewidth,height=3.4cm]{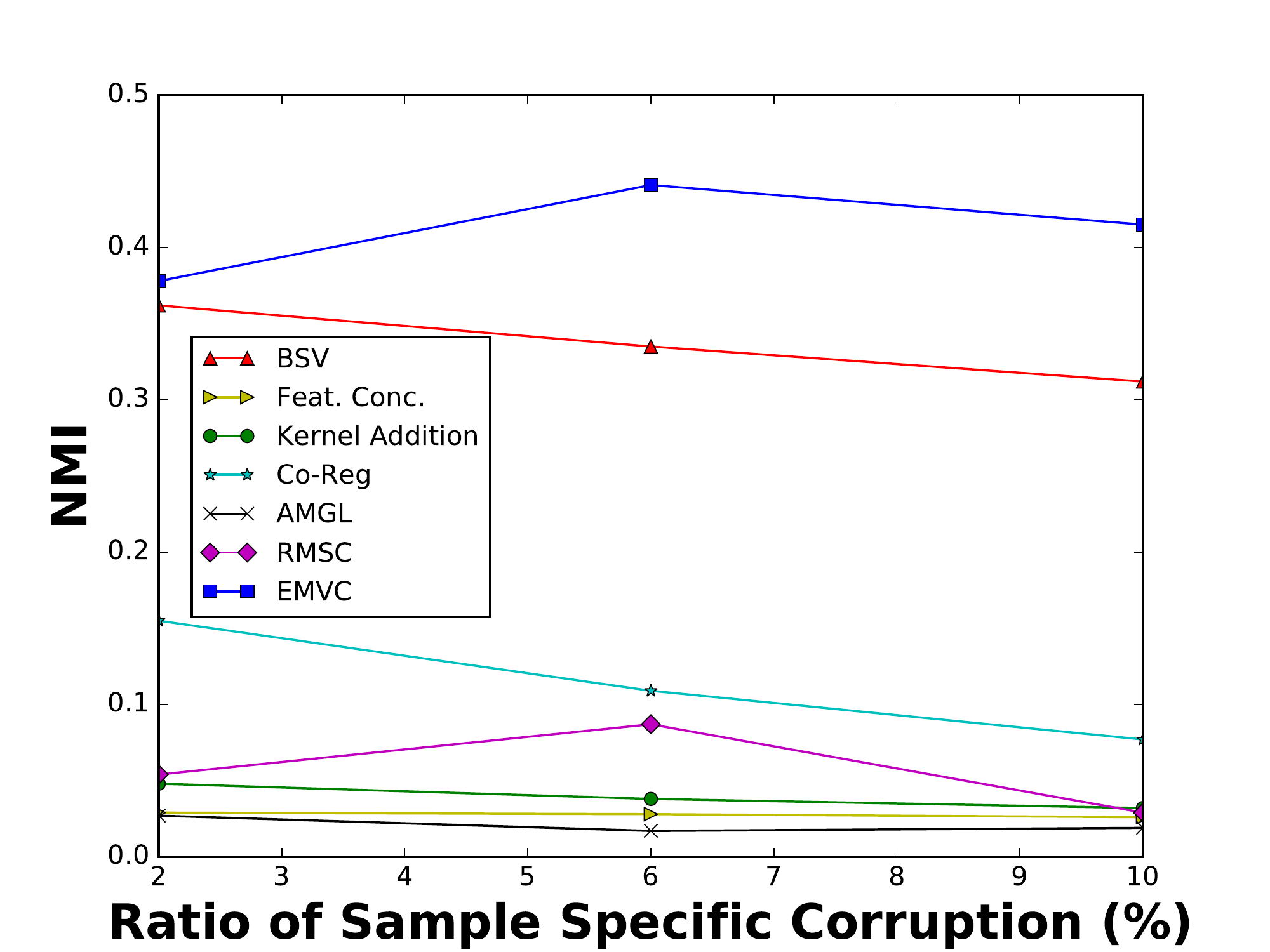}
                \label{fig:CNN-nmi}
        }\\
    \label{fig:error}
    \caption{Clustering performance of erroneous real-world datasets}
 \end{figure*}

\subsection{Experiments on Erroneous Real-World Datasets}
To evaluate robustness of the proposed EMVC method on noise and random corruptions, we add \textit{white Gaussian noise} with different signal-to-noise ratio \{$0.01$, $0.1$, $1$, $10$, $100$\} on FOX (denoted as NRC-FOX) and CNN (denoted as NRC-CNN).
Fig. 4 shows clustering performance of the methods with various signal-to-noise ratio on the contaminated datasets. We can see that EMVC consistently achieves superior performance over the baselines (we only show the results for full version of EMVC, which is superior to its degenerative versions). This observation demonstrates that EMVC is robust against random noise and corruptions.





We investigate robustness of the proposed EMVC method against sample-specific corruptions on FOX (denoted as SSC-FOX) and CNN (denoted as SSC-CNN). 
For this experiment, we randomly select a small portion of samples (2\%, 6\% and 10\%) and replace their feature values in all views by random values. This setting is similar to generation of attribute outliers in \cite{zhao2015dual}. Fig. 4 shows clustering performance of the methods on sample-specific corruptions. The proposed EMVC method outperforms the baselines against this type of error (we only show the results for full version of EMVC, which is superior to its degenerative versions). This is mainly because of $\ell_{2,1}$ norm in our objective function, which has sparse row supports.

\subsection{Hyperparameter Analysis}
To explore the effects of the hyperparameters on the performance, we run experiments on real-world datasets with different values for $\lambda \in \{10^{-9}, 10^{-8}, ..., 10^{8}, 10^{9}\}$ and $\beta \in \{10^{-9}, 10^{-8}, ..., 10^{8}, 10^{9}\}$ and report the average accuracy in Fig. 5. In this Figure, each grid with different shades of colors reflects the clustering quality, where yellow means excellent quality. 
We can see that the performance is fairly stable. EMVC enjoys more promising results when $\lambda > 10^{-3}$ and $\beta > 10^{-3}$, while it is almost insensitive to the hyperparameters in that range. 

\begin{figure*}[h]
        \centering
        \subfloat[WebKB]{
                \centering
                \includegraphics[width=0.20\linewidth,height=2.5cm]{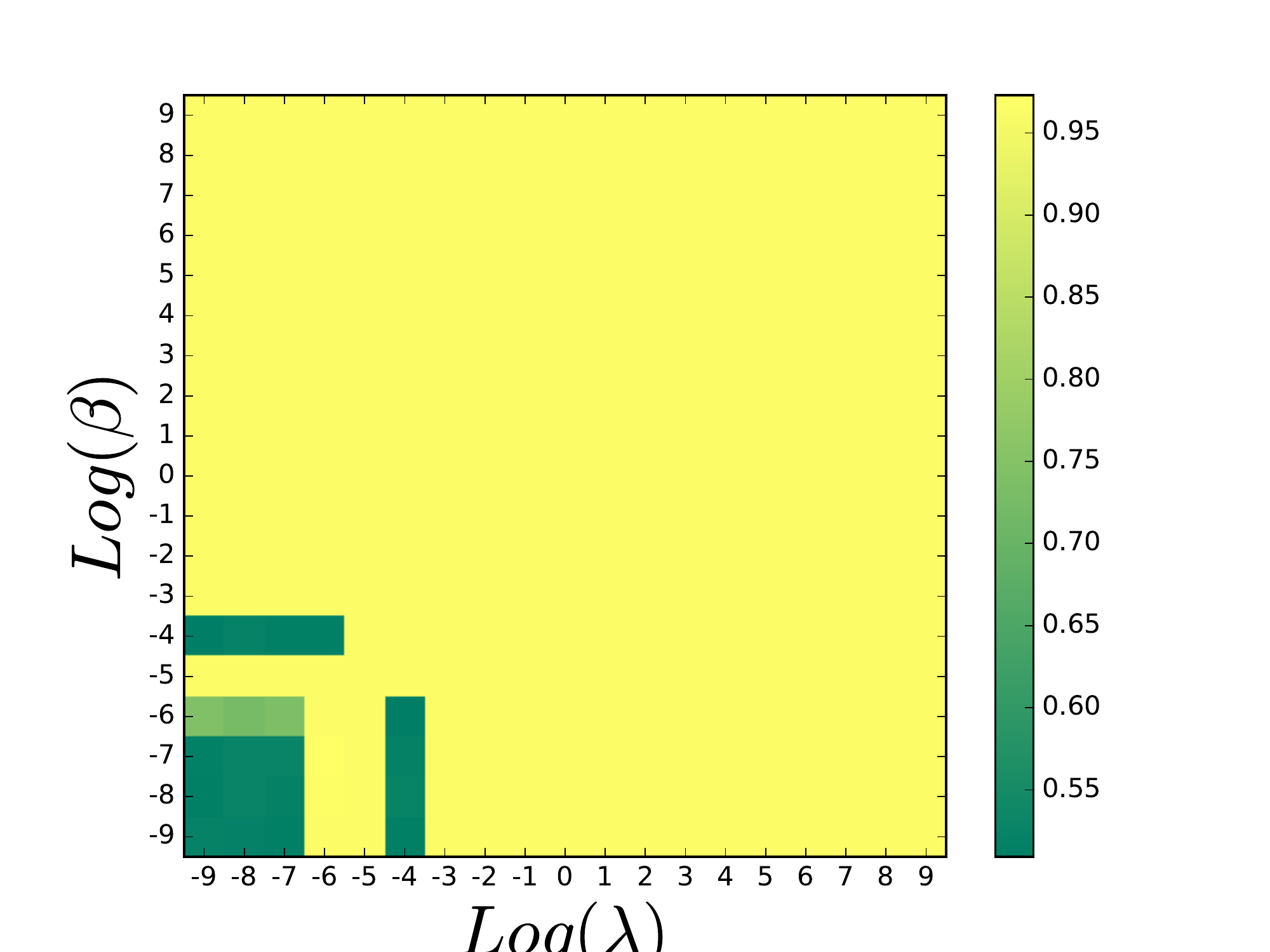}
                \label{fig:Webkb-hyper}\hfill}
        \subfloat[FOX]{
                \centering
                \includegraphics[width=0.20\linewidth,height=2.5cm]{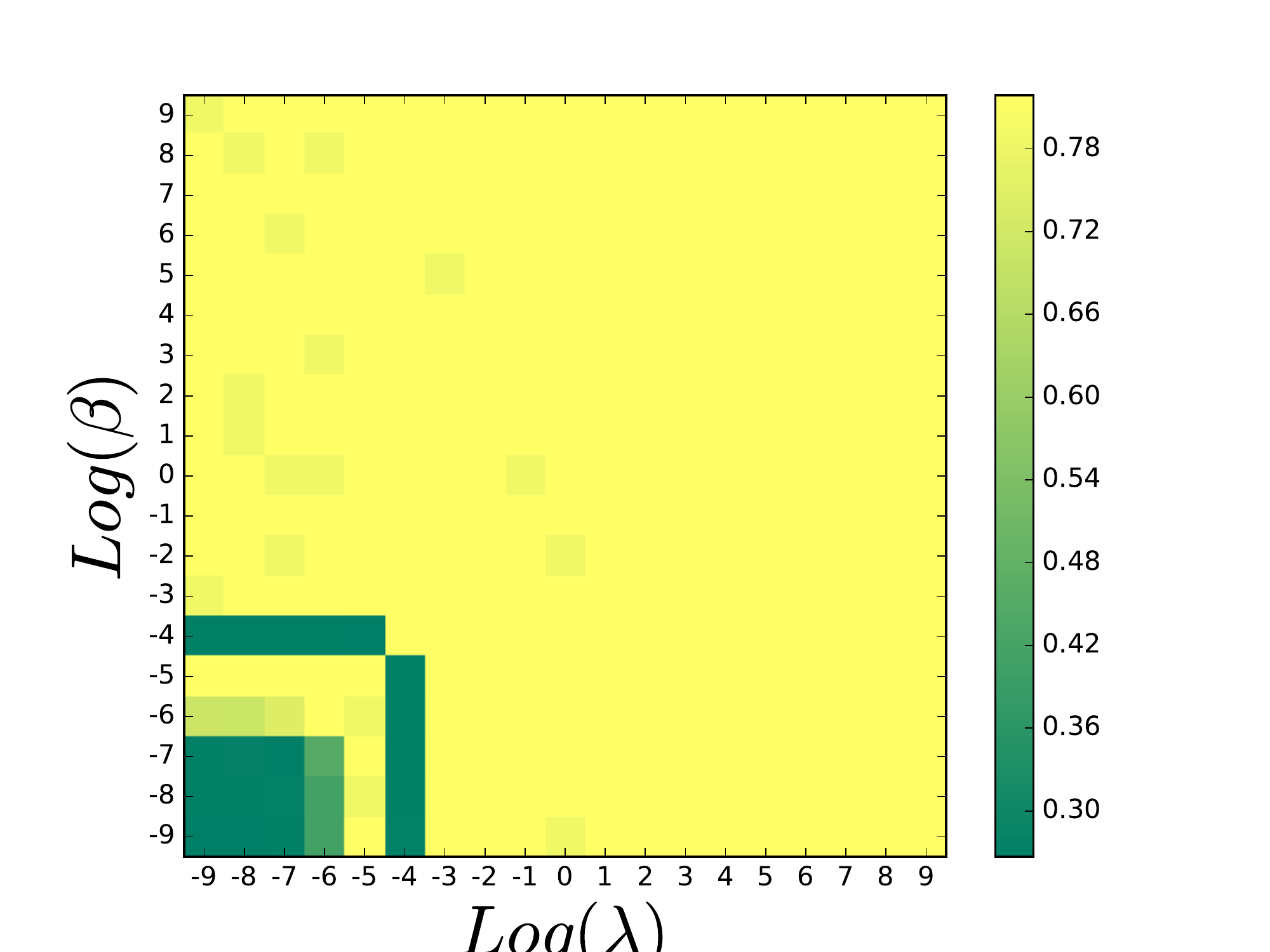}
                \label{fig:FOX-hyper}\hfill}   
        \subfloat[CNN]{
                \centering
                \includegraphics[width=0.20\linewidth,height=2.5cm]{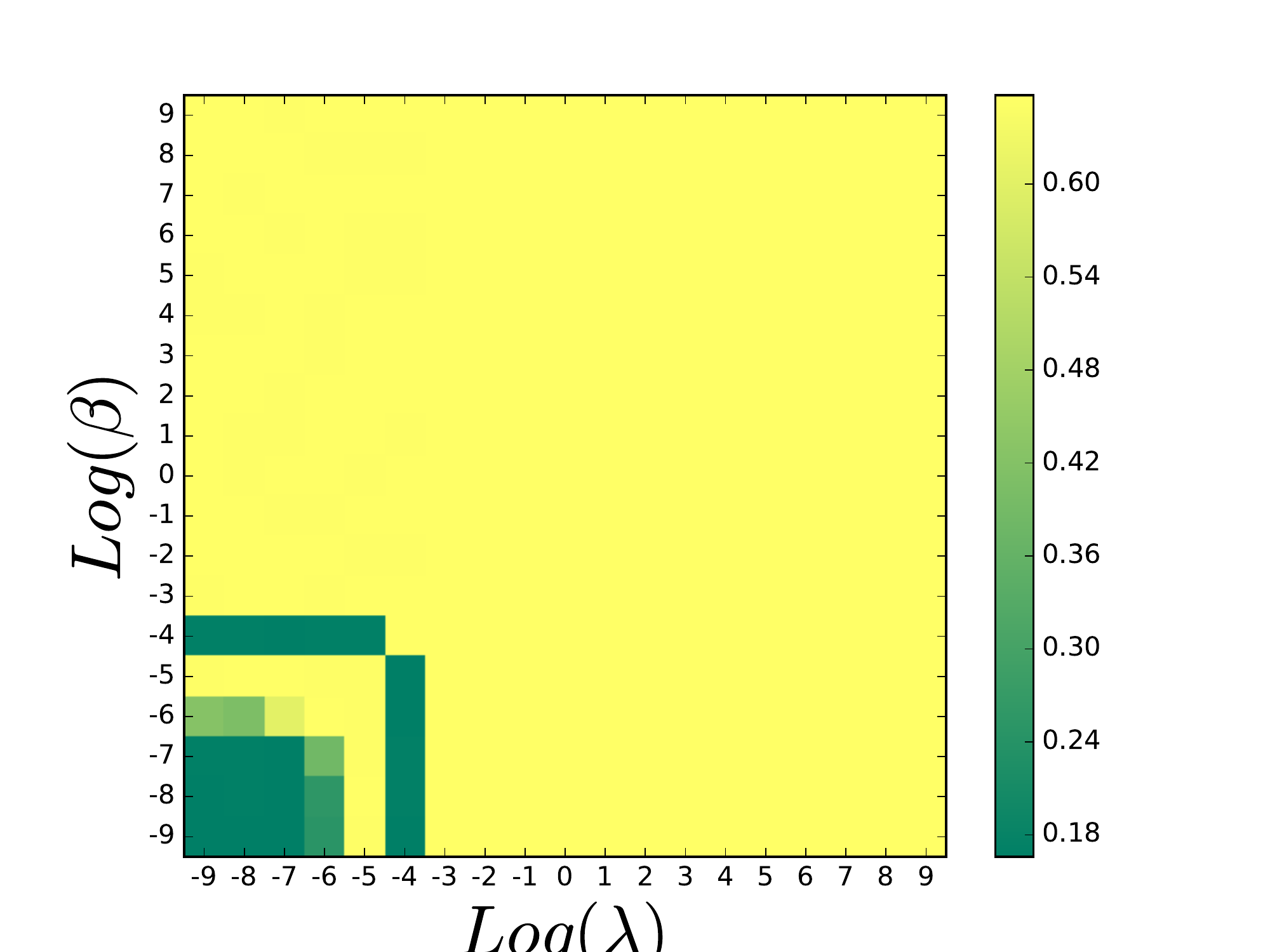}
                \label{fig:CNN-hyper}\hfill}   
        \subfloat[Citeseer]{
                \centering
                \includegraphics[width=0.20\linewidth,height=2.5cm]{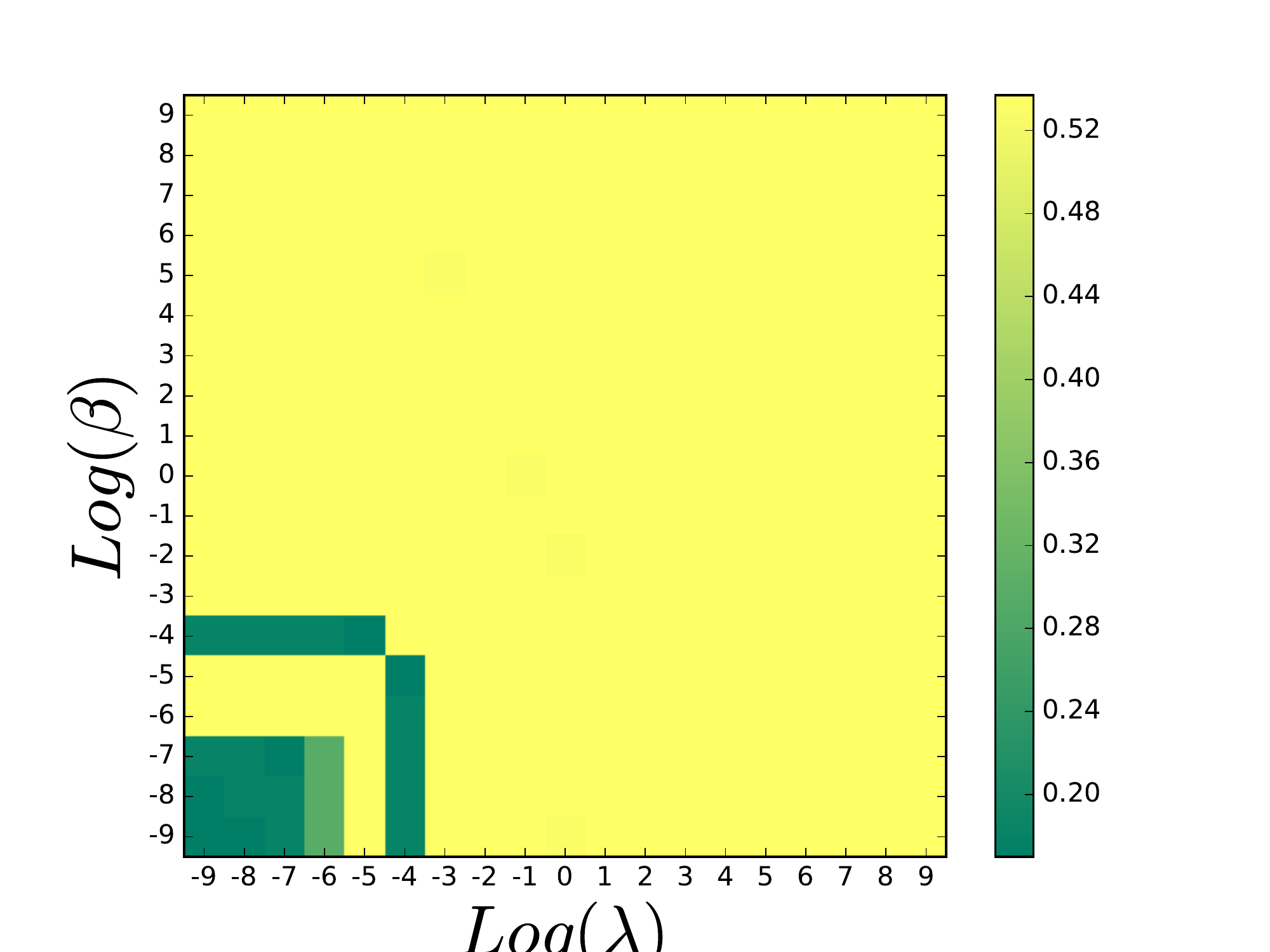}
                \label{fig:CNN-hyper}\hfill} 
        \subfloat[CCV]{
                \centering
                \includegraphics[width=0.20\linewidth,height=2.5cm]{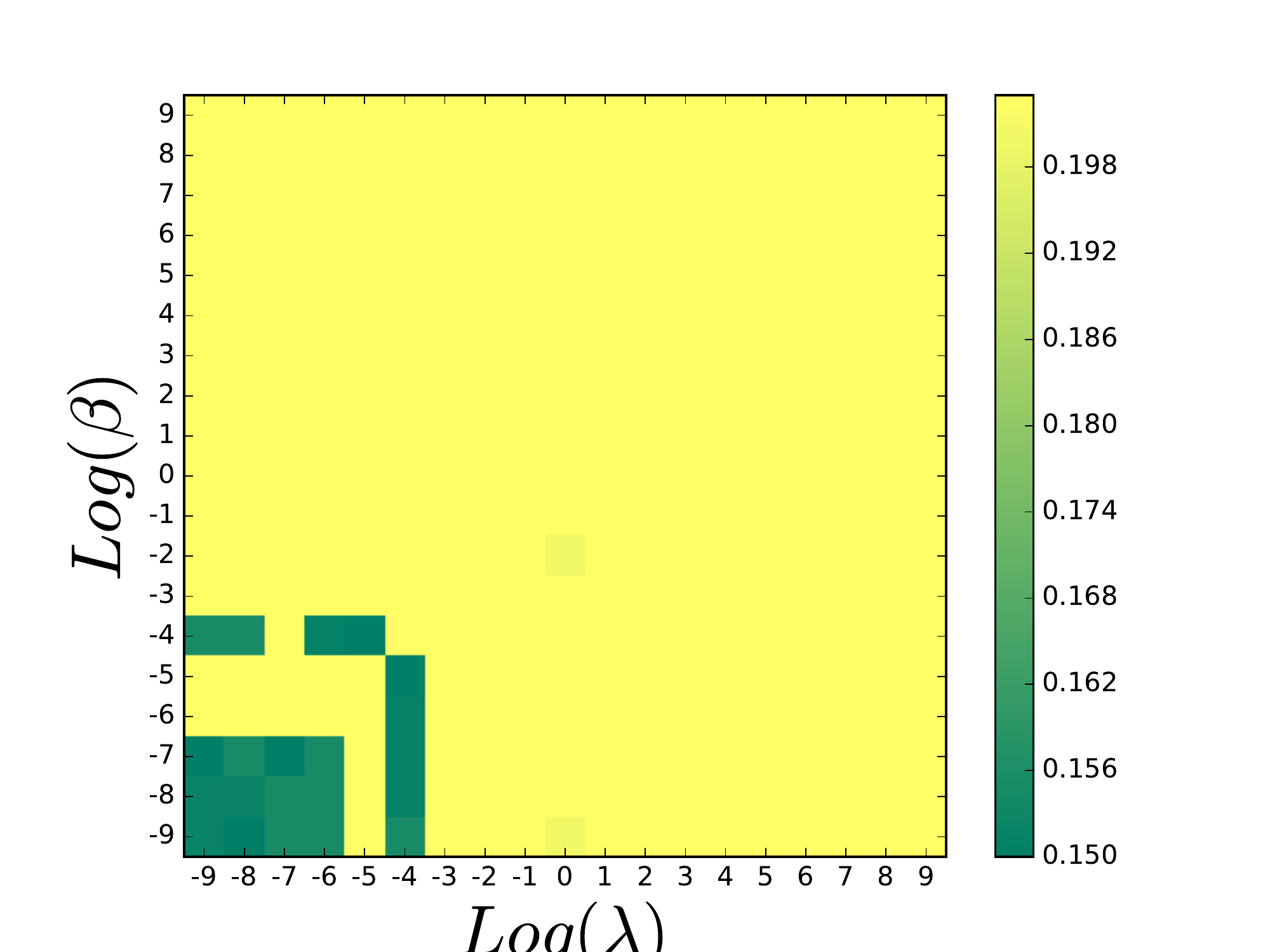}
                \label{fig:CNN-hyper}\hfill}  
    \caption{Sensitivity analysis of regularization hyperparameters (Accuracy)}
 \end{figure*}
 
 \vspace{-0.2cm}

\section{Related Work}
Existing methods for multi-view clustering can be classified into two categories: 1) centralized approaches; 2) distributed approaches \cite{doi:10.1137/1.9781611972788.74}. 
The centralized approaches constructs a new shared representation (i.e., common consensus) across all views 
\cite{Bickel:2004:MC:1032649.1033432,Zhou:2007:SCT:1273496.1273642,Kumar:2011:CMS:2986459.2986617,Xia:2014:RMS:2892753.2892850,Nie:2016:PAM:3060832.3060884}. For example, Bickel and Scheffer presented an algorithm that interchanges the cluster information among different views \cite{Bickel:2004:MC:1032649.1033432}. 
Xia et al. proposed a multi-view clustering method, named as RMSC, that recovers shared transition probability matrix, in favor of low-rank and $\ell_{1}$ regularization \cite{Xia:2014:RMS:2892753.2892850}. 
The proposed EMVC method belongs to this category. Different from RMSC, EMVC builds a shared transition probability matrix by integrating decomposition and group $\ell_{1}$ and $\ell_{2,1}$ norms. EMVC also handles typical error types well.

\vspace{-0.15cm}

The distributed approaches often build separate learners for each individual view and use the information in each learner to apply constraints on other views
\cite{Kumar:2011:CMS:2986459.2986617}. Kumar et al., proposed an approach to combine graphs of each individual view by pairwise co-regularization to achieve better clustering solution \cite{Kumar:2011:CMS:2986459.2986617}. EMVC differs from this category of approaches in a way that it does not construct separate learners. Instead, it recovers a shared transition probability matrix across all views.

\section{Conclusion}
In this paper, we developed a Markov chains method named EMVC for multi-view clustering via a low rank decomposition and two regularization terms. EMVC has several advantages over existing multi-view clustering methods. First, it handles typical types of error well. Second, an iterative optimization framework is proposed for EMVC which is proved to converge.
Compared to the existing state-of-the-art multi-view clustering approaches, EMVC showed better performance on five real-world datasets.

\vspace{-0.2cm}
\balance
\bibliographystyle{IEEEtran}
\bibliography{IEEEabrv,references.bib}

\end{document}